\documentclass[10pt,twocolumn,letterpaper]{article}

\usepackage{cvpr}              

\usepackage[accsupp]{axessibility}

\definecolor{cvprblue}{rgb}{0.21,0.49,0.74}
\usepackage[pagebackref,breaklinks,colorlinks,allcolors=cvprblue]{hyperref}
\usepackage{tcolorbox}
\usepackage{graphicx}
\usepackage{booktabs}
\usepackage{multirow}
\usepackage{makecell}
\usepackage{booktabs}
\usepackage{rotating}
\usepackage{colortbl}
\usepackage{xcolor}
\usepackage{arydshln}
\usepackage{soul}
\usepackage{multirow, array, booktabs}
\usepackage{caption}
\usepackage{subcaption}
\usepackage{xfrac}
\usepackage{graphicx}
\usepackage{calc}
\usepackage{tikz}
\usepackage{rotate}
\usepackage{colortbl}

\def\paperID{2241} 
\def\confName{CVPR}
\def\confYear{2025}

\title{Semantic and Expressive Variation in Image Captions Across Languages}

\author{Andre Ye$^1$, Sebastin Santy$^1$, Jena D. Hwang$^2$, Amy X. Zhang$^{1, 2}$, Ranjay Krishna$^{1, 2}$\\
$^1$University of Washington, $^2$Allen Institute for AI \\
Seattle, Washington \\
{\tt\small andreye@uw.edu},
{\tt\small ssanty@cs.washington.edu},
{\tt\small axz@cs.uw.edu},
{\tt\small ranjay@cs.washington.edu}
}

\begin{document}

\newcommand\andre[1]{{\color{brown}[{#1}]$_{-\text{Andre}}$}}
\newcommand\seb[1]{{\color{orange}[{#1}]$_{-\text{Seb}}$}}
\newcommand\rk[1]{{\color{blue}[{#1}]$_{-\text{Ranjay}}$}}
\newcommand\matt[1]{{\color{green}[{#1}]$_{-\text{matt}}$}}
\newcommand\attention[1]{{}}
\newcommand*\rot{\rotatebox{90}}

\definecolor{lightgray}{gray}{0.9}
\definecolor{gold}{HTML}{ffccbc}
\definecolor{silver}{HTML}{ffecb3}


\newcommand{\first}[1]{\tcbox[on line,colframe=gold,boxsep=0pt,left=0.5pt,right=0.5pt, enlarge top by=0.005cm, enlarge bottom by=0.005cm, arc=0.4pt, top=0.3pt,bottom=0.3pt,colback=gold]{#1}}

\newcommand{\second}[1]{\tcbox[on line,colframe=silver,boxsep=0pt,left=0.7pt,right=0.7pt, enlarge top by=0.005cm, enlarge bottom by=0.005cm, arc=0.4pt, top=0.3pt,bottom=0.3pt,colback=silver]{#1}}

\newcommand\one[1]{{#1}}
\newcommand\two[1]{{#1}}
\newcommand\three[1]{{#1}}


\newcommand{\circled}[1]{%
  \tikz[baseline=(char.base)]{
    \node[shape=circle,draw,inner sep=2pt] (char) {#1};}}

\maketitle

\begin{figure*}[!h]
    \centering
    \includegraphics[width=0.8\textwidth]{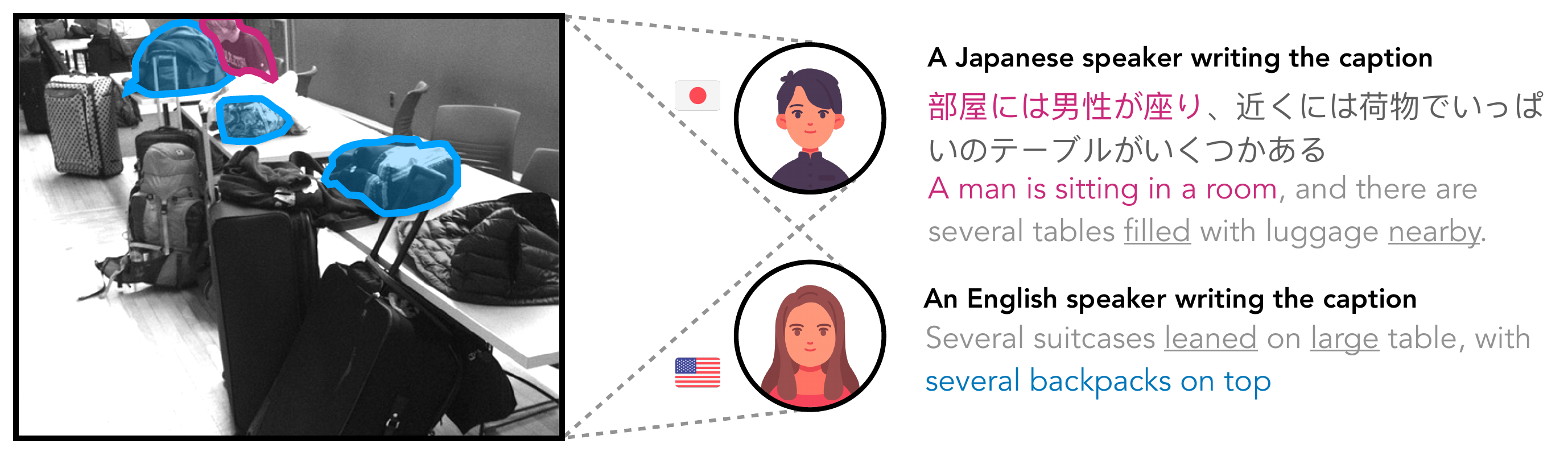}
    \caption{People speaking different languages may caption images differently, noticing and emphasizing different aspects of the image. These examples are drawn from our user study. In this paper, we demonstrate that there are \textit{distributional differences} between the concepts represented in different languages, in addition to the \textit{general variation} in annotator subjectivity/noise. Illustrative example.
    }
    \label{fig:intro-fig}
\end{figure*}

\begin{abstract}
    
Most vision-language models today are primarily trained on English image-text pairs, with non-English pairs often filtered out. 
Evidence from cross-cultural psychology suggests that this approach will bias models against perceptual modes exhibited by people who speak other (non-English) languages.
We investigate semantic and expressive variation in image captions across different languages; we analyze both human-annotated datasets and model-produced captions. 
By analyzing captions across seven languages (English, French, German, Russian, Chinese, Japanese, Korean) in high-quality image captioning datasets (Crossmodal and Visual Genome), we find that multilingual caption sets tend to provide richer visual descriptions than monolingual (including English-only) ones; multilingual sets contain 
$46.0\%$ more objects, 
$66.1\%$ more relationships, and 
$66.8\%$ more attributes.
We observe the same results with multilingual captions produced by LLaVA and the Google Vertex API: for example, compared to monolingual captions, they cover
$21.9\%$ more objects, 
$18.8\%$ more relations, and 
$20.1\%$ more attributes. 
These suggest that, across a large number of samples, different languages bias people and models to focus on different visual concepts.
Finally, we show that models trained on image-text data in one language perform distinctly better on that language's test set.
Our work points towards the potential value of training vision models on multilingual data sources to widen the range/variation of descriptive information those models are exposed to.
    
\end{abstract}

\section{Introduction}


Typically, vision-language models are trained with large quantities of image-text pairs scraped from the web~\cite{Thomee2015TheND,schuhmann2022laionb,Erdem2022NeuralNL}.
To obtain high quality image-text pairs, these web datasets are filtered with models like CLIP~\cite{Radford2021LearningTV, ilharco_gabriel_2021_5143773} with the goal of ensuring that each text sample is sufficiently descriptive of or related to the corresponding image.
However, this filtering tends to be biased towards English text samples, such that the majority of samples in popular image-text datasets are in English~\cite{nguyen2024multilingualdiversityimprovesvisionlanguage, hong2024whoswhosoutcase,schuhmann2022laionb}.
Many vision-language models are even trained only on English~\cite{Radford2021LearningTV, tan-bansal-2019-lxmert, li2022blip, kim2021viltvisionandlanguagetransformerconvolution}, excluding other languages entirely.

Prior work shows that models tend to fail at tasks where naturally occurring text is uncommon, such as negation~\citep{Dobreva2021InvestigatingNI}.
Therefore, increasing the lexical diversity of naturally occurring text is important: it allows models to experience a broader range of linguistic patterns that might be underrepresented within a certain set of scraped text, while maintaining the scale that is only possible through training on naturally occurring data.
Thus, biasing vision datasets towards English-only image-text pairs may risk excluding the equally relevant and possibly unique visual concepts produced in languages other than English.
Indeed, a wide body of work from the cross-cultural psychology suggests that \textit{people from across the world, speaking different languages and living in different cultural contexts, tend to describe the visual world differently}.
Early work in psychology suggests that fundamental aspects of visual perception, such as perception of length~\cite{Segall1967TheIO}, geometrical intuition~\cite{Pedersen1983TheMI, Dawson1973DevelopmentalIO}, and depth~\cite{Jahoda1974PictorialDP} vary across people from different geographic backgrounds.
More recent work has shown that individuals from different cultural backgrounds exhibit differences in how they look at, understand, and talk about visual scenes~\cite{Nisbett2003CultureAP,Koo2018AnalyticVH,enk2020CrossculturalDI}.
For instance, when looking at the same image, Americans tend to describe the focal objects and its attributes, whereas Japanese tend to focus on the relations between objects \cite{Masuda2001AttendingHV}.

Drawing on this work, we hypothesize the content distribution of visual descriptions varies by the structure and use of specific languages  -- and that these differences are detectable in computer vision datasets and model behaviors. 
This hypothesis is further supported by observed differences in spoken language: German's complex morphosyntactic system provides events with nuanced understanding of spatial relationships~\cite{prange2021draw};
Russian verbs of motion require speakers to specify directionality~\cite{10.1111/modl.12177}; English uses the generic verb of motion ``to go'' whereas a Russian speaker must specify whether the motion has a destination/purpose, the mode of transportation, and other possibly visually salient variables.
Similar factors associated with different languages might encourage the expression of different visual concepts in descriptions.
If true, our hypothesis suggests that models trained on English-only data would be further improved if exposed to a more global view.

In this paper, we investigate \textbf{whether captions produced by people (in datasets) and by models tend to vary semantically and expressively across different languages}. 
In our context, \textit{semantics} refers to a caption's content (what a caption ``says'') as represented by scene graphs (objects, attributes, relations)~\cite{Johnson2015ImageRU, Krishna2016VisualGC} and \textit{expression} refers to how its content is communicated as measured by linguistic measures such as concreteness, authenticity, and tone~\cite{Boyd2022LIWC22}.
We measure the variation of semantic and expression measurements between multilingual and monolingual sets of captions.
If we systematically observe a greater variation in multilingual caption sets vs. monolingual sets (for example, if scene graphs created from multilingual caption sets are larger than monolingual ones), then this demonstrates that captions have different distributions of concepts in different languages.
We analyze the human-produced captions from the high-quality Crossmodal dataset across seven diverse languages (English, French, German, Russian, Chinese, Japanese, Korean).
The Crossmodal dataset keeps annotator instructions consistent across languages, ensuring fair comparison of multilingual and monolingual caption sets in our analysis by eliminating possible confounding sources of variation across languages.

Our results demonstrate that multilingual caption sets demonstrate higher semantic and expression variation than monolingual caption sets.
Importantly, while it is known that different people and models produce fairly widely varying captions \textit{in general}~\cite{hutchinson2022underspecification, Blandfort2017ImageCI}, our claim is that there are also \textit{broad distributional differences} between captions produced in different languages, when evaluated over large numbers of samples.
Multilingual scene graphs are larger overall (that is, they cover more content) than monolingual scene graphs, with an increase of 46.0$\%$ objects, 66.1$\%$ relationships, and 66.8$\%$ attributes compared to scene graphs built from English-only captions.
Multilingual caption sets also exhibit broader expressive variation, such as an increase of 53.4$\%$ in range of tonality and 42.1$\%$ in coverage of the embedding space. 
We extend this analysis to vision-language model behaviors (LLaVA and the Vertex API \verb|imagetext-001|), and observe similar patterns in model-generated captions: multilingual scene graphs are larger than monolingual scene graphs overall by  
21.9$\%$ objects, 
18.8$\%$ relations, 
and 20.1$\%$ attributes, 
and multilingual caption sets have
63.0$\%$ wider range of tonality 
and nearly 92.4$\%$ wider coverage of the embedding space.

We further analyze how finetuning on linguistically diverse data affects a model's captioning capabilities.
Models are often finetuned on captions produced by humans (in datasets) or by other models (distillation).
To understand this, we fine-tune models on captions from one language and evaluate them on captions from another language.
Models finetuned on language $X$ perform significantly better on the test set from language $X$, suggesting that models internalize language-specific distributional characteristics.
For example, a model finetuned on captions translated from Japanese attains a SPICE F-score of .27 on reference captions translated from Japanese, but only .23 from English.


In sum, our \textbf{primary contribution} is providing evidence that captions produced by people and models vary by language.
To do so, we give specific measurement methods on which multilingual distributions of captions have wider variation in content or information than monolingual distributions.
Our work may help reframe the ``curse of multilinguality'' by emphasizing the diverse range of visual concepts in multilingual datasets.
We are limited by our specific seven languages and mid-sized scale, but we believe our study provides a focused demonstration of the properties of multilingual data.
Our work provides one possible explanation to findings in recent work~\cite{nguyen2024multilingualdiversityimprovesvisionlanguage} --- that training on multilingual data at large scale improves vision representations --- by understanding how the information in multilingual caption distributions might differ from monolingual ones, and relatedly, how monolingual captions differ across languages.
See \S\ref{related-work} for related work and \S\ref{limitations} for limitations.

\section{Measuring variation in datasets}
\label{sec:dataset-variation}

We want to understand how captions in computer vision \textit{datasets} vary across languages.
Variation across languages in datasets may have downstream effects in associated benchmarks and downstream models.
We choose to perform a close analysis of the Crossmodal (``XM'') dataset~\cite{Thapliyal2022Crossmodal3600AM} because of its high quality and well-designed data annotation procedures.
XM contains image descriptions in 36 languages over 3.6k geographically diverse images.
To eliminate potential bias in captioning instructions across annotators (e.g., varying instructions in different languages), all annotators are provided with the same annotation instruction and required to be reading-proficient in English (in addition to being native/fluent in the target language).
To ensure consistency in caption style, annotators across languages are primed with the same base caption.
These procedures ensure that the XM dataset serves as a prime exemplar to fairly study differences between captions produced in different languages for the same sets of images.

Throughout the paper, we analyze captions produced in seven languages: English, French, German, Russian, Chinese, Japanese, and Korean.
Together, these 7 languages encompass a wide range of typologically diverse linguistic families.
Moreover, the speakers of these languages originate from a large variety of cultures and experiences.

\subsection{Translation for fair comparison}
\label{translation_fair_comparison}

We translate all captions into English with GPT-4~\cite{OpenAI2023GPT4TR} 
(\verb|gpt-4-0613|)
to ensure fair comparison of caption content across languages by eliminating linguistic confounders.
Our measurements of semantic (\S\ref{subsec:data-semvar}) and expressive (\S\ref{subsec:data-exprvar}) variation use tools like parsers, embedding models, and tokenizers which are either language-specific (e.g., entirely separate parsers for different languages~\cite{rafferty-manning-2008-parsing, chang-etal-2009-discriminative}) or language-biased (e.g., multilingual embeddings encode language-specific information~\cite{libovický2020language, otmakhova-etal-2022-cross, chang2022geometry, philippy2023identifying}).
When all captions are translated into English, we can analyze on the deeper differences in caption content and expression on ``common ground'', without these linguistic confounders.

We chose GPT-4 because we can specify the manner and procedure for translation in the prompt to a degree not available for may other models; specifically, in our case it was more important for translations to be exact and preserve all factual details than to adhere to other translation values and metrics~\cite{celikyilmaz2021evaluation}.
Several works have demonstrated GPT-4’s competence and control on multilingual tasks~\cite{raunak2023leveraging, lee2023straw, jiao2023chatgpt, manakhimova-etal-2023-linguistically, zhu2024multilingualmachinetranslationlarge}, particularly in capturing nuances of expression and meaning specific to a language.

We conduct a \textbf{human evaluation} to verify that the translations preserve salient conceptual information. 
We recruit 15 native speakers to rate 30 translations across each of the 6 non-English languages.
Native speakers describe the translations  as near perfect ($\mu = 4.68$ on a 1-5 scale, where 5 is ``perfect'' and 4 is ``most information preserved'').
Importantly, participants considered $98.42\%$ of important information in a visual scene (objects, spatial relations, color, etc.) as faithfully preserved across translation. 
See Appendix~\ref{translation_quality_check} for more details.
We also verify statistically that variation among translated captions is not explained by the original language of the captions in Section~\ref{finetuning}.
Participants were allowed to give open-ended feedback in the user study, and we find that even the most extreme corrections were minor -- e.g. replacing ``on'' with ``above'' and ``memories'' with ``impressions'' -- and therefore do not noticeably alter measurements.
Therefore, we believe the translations did not introduce significant analysis-biasing artifacts.

Nonetheless, we repeat the representation variance experiment described in \S\ref{subsec:data-exprvar} and \S\ref{subsec:expvar-lang} with multilingual embeddings \textit{without translation} and find that the same result holds as described (using English embeddings with translation) (see \S\ref{multilingual_embedding}).
This shows that our results still hold when using tools that can be theoretically applied across languages (even though they are language-biased).
We describe all of our experiments in the main paper under the assumption of English translation for fair comparison.


\subsection{Semantic variation across languages}
\label{subsec:data-semvar}

We want to understand how captions across languages vary in their \textit{semantic content}.
A caption's semantic content is what it says about the objects in the image and how they relate to each other.
We ask: \textit{do captions across languages mention the same objects and relations, or are there systematic differences?}
Scene graphs are a common representation for this kind of information~\cite{Johnson2015ImageRU,Choi2022SceneGP,ji2020action}.

\begin{table*}[t]
\footnotesize
\caption{\textbf{XM (dataset) evaluation results.} Results compare monolingual (e.g., all-English) against multilingual (e.g., English, French, Chinese) caption sets against both semantic (``Sem.'') and selected expressive (``Expr.'') measurements. The multilingual ``avg'' column indicates the average of all multilingual language three-tuples, since not all can be displayed. \textbf{Observe that multilingual measurements are almost always larger than monolingual ones.} See all data aggregated at Table~\ref{primary-table}.}
\centering
\begin{tabular}{cc|rrrrrrr|>{\columncolor{gray!25}}r|rrr|>{\columncolor{gray!25}}r}
\toprule
& & \multicolumn{8}{c|}{Monolingual} & \multicolumn{4}{c}{Multilingual} \\
& Metric & en & fr & de & ru & zh & ja & ko & avg & en-fr-zh & fr-zh-ru & de-fr-ru & avg \\
\midrule
\multirow{3}{*}{\rotatebox{90}{Sem.}} 
& Objects & 2.59 & 2.92 & 3.16 & 3.03 & 2.99 & 3.41 & 2.71 & 2.98 & 3.71 & 3.92 & 3.93 & 4.35 \\
& Relations & 1.54 & 1.76 & 1.94 & 1.88 & 1.71 & 1.99 & 1.59 & 1.77 & 2.41 & 2.57 & 2.57 & 2.94 \\
& Attributes & 1.27 & 1.66 & 1.97 & 1.74 & 2.01 & 2.47 & 1.46 & 1.78 & 2.36 & 2.59 & 2.55 & 2.97 \\
\midrule
\multirow{4}{*}{\rotatebox{90}{Expr.}}
& Concreteness & 1.80 & 2.24 & 2.11 & 2.03 & 2.26 & 2.25 & 2.14 & 2.12 & 2.08 & 2.13 & 2.10 & 2.17 \\
& Authenticity & 33.21 & 40.09 & 42.22 & 32.35 & 34.02 & 33.83 & 35.28 & 35.86 & 53.12 & 54.41 & 53.35 & 53.21 \\
& Tone & 8.62 & 12.16 & 9.74 & 10.90 & 10.59 & 9.18 & 9.07 & 10.04 & 13.78 & 14.69 & 15.42 & 15.40 \\
& Embeddings & .38 & .42 & .43 & .40 & .46 & .42 & .43 & .42 & .54 & .54 & .49 & .52 \\
\bottomrule
\end{tabular}
\label{tab:xm_combined_evaluation}
\end{table*}

\noindent\textbf{Method:}
Formally, a scene graph $\mathcal{G}$ is defined as a list of tuples of the form $(\text{object},  \text{attribute})$ or $(\text{subject}, \text{predicate}, \text{object})$.
Let $|\mathcal{G}|_{\text{obj}}$, $|\mathcal{G}|_{\text{rel}}$, and $|\mathcal{G}|_{\text{attr}}$ be the unique number of objects, relations, and attributes in a scene graph $\mathcal{G}$, respectively.
These give measurements for the `size' of $\mathcal{G}$.
Let $\textsf{SG}(c)$ construct a scene graph from a caption $c$, and $\mathcal{G}_1 \cup \mathcal{G}_2$ represent the ``union'' of scene graphs --- a new scene graph with shared content (objects, relations, attributes) only counted once.
For each image $i$ out of all images $I$ in the dataset, there is a corresponding set of captions $C_i$, where each caption may be denoted by $c^{l}_{i,k}$, where $l$ is a language and $k$ is the caption index for captions in that language. 
Let $\mathcal{L} = \{ \text{en}, \text{fr}, \text{de}, \text{ru}, \text{zh}, \text{ja}, \text{ko} \}$ be the set of languages.
We create a monolingual set of captions for language $l$: $\text{mono}_i^l = \{ c^l_{i,1}, \hdots, c^l_{i,K} \}$ and a multilingual set of captions for a set of languages $L$: $\text{multi}_i^L = \{ c^{L_1}_{i,1}, \hdots, c^{L_K}_{i,1} \}$.
Let $M(\mathcal{G}) \in \{ |\mathcal{G}|_{\text{obj}}, |\mathcal{G}|_{\text{adj}}, |\mathcal{G}|_{\text{attr}} \}$ be a metric for scene graph size.
We want to compare $\mathbb{E}_{i \in I, l \in \mathcal{L}} [M(\bigcup \textsf{SG}(\text{mono}_i^l))]$ (mean size of scene graphs from monolingual captions) against $\mathbb{E}_{i \in I, L \subset \mathcal{L}; |L|=K} [M(\bigcup_{l \in L} \textsf{SG}(\text{multi}_i^L))]$ (mean size of scene graphs from multilingual captions).
If the latter is significantly larger than the latter, then we conclude that, there tend to be distributional non-overlaps/differences across the concepts covered in captions across languages (apart from differences across captions due to general annotator subjectivity).
To implement $\textsc{SG}(c)$ which constructs a scene graph from a caption $c$, we parse translated XM captions into scene graphs using FLAN-T5~\cite{Chung2022ScalingIL} model fine-tuned on the FACTUAL-MR dataset.
We find that LLM-based parsers were better able to resolve complex semantic relationships which often arose in the descriptions than conventional syntax-based parsers~\cite{schuster2015generating}.
To implement the union of scene graphs, we concatenate their list representations and remove double counting of shared concepts (objects, relations, or attributes).
We canonicalize the concepts using WordNet path similarity~\cite{Miller1995WordNetAL} and cosine similarity between concept embeddings~\cite{reimers-2019-sentence-bert} to merge concepts which have different text but refer to the same concept, erring on the side of merging.
Multiple relationships between two entities are merged into a single edge.

\noindent\textbf{Results:} We find that \textbf{multilingual unioned scene graphs cover more objects, relations, and attributes than monolingual ones}~(Table~\ref{tab:xm_combined_evaluation}).
For instance, unioning the scene graph from an English caption with the scene graphs from a French and a Chinese caption increase the resulting scene graph by
$46.0\%$ objects, 
$66.1\%$ relations, 
and $66.8\%$ attributes. 
Further computing the sizes of intersections \textit{between} \textit{monolingual} unioned scene graphs 
shows that any two languages share only $63.1\%$ of objects and $39.5\%$ of relations in common on average (Table~\ref{tab:crossling_metrics}).
Since image captions are short and targeted to focus on salient parts of the image, these observed differences between languages likely indicate different tendencies in what counts as salient across caption sets written by humans who speak different languages.
Qualitatively, we observe that many of the objects represented in multilingual scene graphs but not in English scene graphs tend to be background objects and objects at varying levels of object composition (e.g. keyboard \textit{and} CTRL key). Moreover, multilingual scene graphs tend to include more information about perspective (e.g. ``zooming'', ``fore/background''), color, size, and other details.
See Figures~\ref{fig:scene-graph_union} and \ref{fig:big-scene-graph-spread} for detailed examples.
Importantly, this demonstrates that there are \textbf{distributional non-overlaps between the concepts represented in captions across languages}.

\subsection{Expressive variation across languages}
\label{subsec:data-exprvar}

We also want to understand how captions across languages vary by \textit{expression}.
Expression refers to \textit{how} concepts are conveyed through language.

\noindent\textbf{Method:} We consider five linguistic measures of expression:
\textit{Concreteness} is computed across noun objects and indicates how much a word refers to a perceptible entity~\cite{Brysbaert2014ConcretenessRF}. For instance, ``purple'' has a high concreteness rating (4.04 on a 5-point scale), whereas ``beautiful'' has a low one (2.16).
Text is \textit{analytic} when it expresses logical and hierarchical thinking patterns,
demonstrates \textit{clout} when it exhibits social status and confidence,
is \textit{authentic} when it is spontaneous,
and varies by \textit{tone}.
These last four measures are provided by the Linguistic Inquiry and Word Counts (LIWC) framework~\cite{Tausczik2010ThePM}, and are computed using word dictionaries and aggregation algorithms informed by linguistic and psychological research~\cite{Boyd2022LIWC22}.
Many previous works have demonstrated the use of these measures in the quantification of emotional and psychological expression in text from a variety of contexts \cite{schaefer2023literature, junghaenel2008linguistic, 10.2308/AJPT-2020-060, Kahn2007MeasuringEE}.
A sixth measure is a model embedding, which captures some semantic but also important expressive dimensions of text~\citep{conneau2018cramsinglevectorprobing, chen2013expressivepowerwordembeddings, artetxe-etal-2018-uncovering}.

To measure expressive variation, we compare the \textit{expressive coverage} of multilingual caption sets and monolingual caption sets, where the expressive coverage of a set of text $\mathcal{T}$ as measured by a measurement $M$ is $C_M(\mathcal{T}) = \max (M(\mathcal{T})) - \min (M(\mathcal{T}))$, or the maximum width of $M$ spanned by elements of $\mathcal{T}$.
Note that because embeddings are not one-dimensional, expressive coverage is measured as the maximum pairwise cosine distance between two points in a set.
Like previously, we compare $\mathbb{E}_{i\in I, l \in \mathcal{L}} [C_M(\text{mono}_i^l)]$ and $\mathbb{E}_{i \in I, L \subset \mathcal{L}} [C_M(\text{multi}_i^L)]$.
Following the interpretation methodology from \S\ref{subsec:data-semvar}, if the multilingual sets have more expressive coverage than monolingual sets, then this demonstrates some non-overlap in the distribution of captions' expressive dimensions (as measured by the aforementioned metrics).


\noindent\textbf{Results:} 
Our results demonstrate \textbf{variance in caption expression across languages}.
We find that across all metrics, expressive coverage is generally higher for multilingual rather than monolingual subsets (Table~\ref{tab:xm_combined_evaluation}).
For example, the range of tonality in caption sets widens by 53.4$\%$ for multilingual sets.
We also find that the mean width of caption sets in model representations is larger for multilingual sets than monolingual sets.
For example, the average coverage in embedding space of English XM captions increases from 0.38 to 0.54 when switching to an English, French, and Chinese caption set.
This means that training with captions from multiple languages as opposed to one language may expose the model to a measurably wider range of caption expressiveness.


% ...
\section{Measuring variation in model outputs}
\label{sec:model-var}

\begin{table*}[t]
\footnotesize
\caption{\textbf{LLaVA and Vertex (model output) evaluation results.} Results compare monolingual (e.g., all-English) against multilingual (e.g., English, French, Chinese) caption sets against both semantic (``Sem.'') and selected expressive (``Expr.'') measurements. The multilingual ``avg'' column indicates the average of all multilingual language three-tuples, since not all can be displayed. \textbf{Observe that multilingual measurements are almost always larger than monolingual ones.} See all data aggregated at Table~\ref{primary-table}.}
\centering
\begin{tabular}{c|cc|rrrrrrr|>{\columncolor{gray!25}}r|rrr>{\columncolor{gray!25}}r}
\toprule
& & & \multicolumn{8}{c|}{Monolingual} & \multicolumn{4}{c}{Multilingual} \\
& Model & Metric & en & fr & de & ru & zh & ja & ko & avg & en-fr-zh & fr-zh-ru & de-fr-ru & avg \\
\midrule
\multirow{7}{*}{\rotatebox{90}{LLaVA}} & \multirow{3}{*}{\rotatebox{90}{Sem.}} 
& Objects & 4.54 & 5.05 & 5.26 & 4.52 & 4.54 & - & - & 4.78 & 6.14 & 6.15 & 6.25 & 5.93 \\
& & Relations & 3.79 & 4.21 & 4.42 & 3.67 & 3.66 & - & - & 3.95 & 4.85 & 4.76 & 4.97 & 4.54 \\
& & Attributes & 2.75 & 3.47 & 3.50 & 2.76 & 3.25 & - & - & 3.15 & 4.00 & 3.99 & 4.07 & 3.86 \\
\cmidrule(lr){2-15}
& \multirow{4}{*}{\rotatebox{90}{Expr.}}
& Concreteness & 2.17 & 2.44 & 2.53 & 2.27 & 2.33 & - & - & 2.35 & 2.54 & 2.54 & 2.57 & 2.56 \\
& & Authenticity & 35.97 & 38.01 & 34.33 & 37.20 & 44.63 & - & - & 38.03 & 54.94 & 54.64 & 54.82 & 53.15 \\
& & Tone & 6.79 & 9.53 & 16.64 & 16.48 & 11.56 & - & - & 12.20 & 16.74 & 20.79 & 19.80 & 16.56 \\
& & Embeddings & .22 & .29 & .28 & .29 & .36 & - & - & .29 & .45 & .47 & .43 & .47 \\
\midrule
\multirow{7}{*}{\rotatebox{90}{Vertex}} & \multirow{3}{*}{\rotatebox{90}{Sem.}} 
& Objects & 3.65 & 3.51 & 3.60 & 3.86 & 3.46 & 3.13 & 3.18 & 3.48 & 4.13 & 4.24 & 4.13 & 4.17 \\
& & Relations & 2.96 & 2.83 & 2.89 & 3.20 & 2.68 & 2.37 & 2.47 & 2.77 & 3.45 & 3.48 & 3.38 & 3.40 \\
& & Attributes & 1.67 & 1.67 & 1.79 & 1.86 & 1.66 & 1.59 & 1.62 & 1.98 & 2.29 & 2.40 & 2.33 & 2.33 \\
\cmidrule(lr){2-15}
& \multirow{4}{*}{\rotatebox{90}{Expr.}}
& Concreteness & 1.64 & 1.64 & 1.67 & 1.66 & 1.51 & 1.50 & 1.56 & 1.60 & 1.75 & 1.74 & 1.73 & 1.81 \\
& & Authenticity & 23.21 & 22.80 & 21.68 & 23.07 & 23.51 & 25.01 & 21.67 & 22.99 & 40.16 & 36.94 & 31.85 & 38.06 \\
& & Tone & 1.85 & 1.84 & 2.03 & 2.63 & 2.05 & 1.96 & 2.05 & 2.06 & 3.98 & 4.10 & 4.19 & 3.92 \\
& & Embeddings & .19 & .18 & .19 & .22 & .20 & .19 & .37 & .22 & .37 & .33 & .38 & .49 \\
\bottomrule
\end{tabular}
\label{tab:llava_vertex_combined_evaluation}
\end{table*}

In \S\ref{sec:dataset-variation}, we demonstrated semantic and expressive variations in the XM dataset.
These results make sense in light of social sciences studies that show how linguistic factors associated with different languages encourage the production of specific kinds of information.
However, we also want to study if multilingual vision models also exhibit similar differences when producing captions in different languages.
We ask the question: ``\textit{what kinds of information differences might there be across captions produced by models in different languages for the same images?}''

We study two models: LLaVA and the Google Vertex API.
LLaVA~\cite{Liu2023VisualIT} is a vision-language model capable of generating multilingual captions. Although LLaVA is trained with English data, it inherits multilingual capabilities from its large language model (LLM) component, LLaMA~\cite{Touvron2023LLaMAOA}. 
In total, we generate 54k captions for 3.6k Crossmodal images in 5 languages with 3 captions each, excluding Japanese and Korean for text quality issues.
LLaVA allows us to study the significance of language factors across descriptions in an isolated fashion, because the vision representations are held constant across text prompts in different languages.
See Appendix~\ref{app:llava-multilingual} for more details on probing multilingual behavior.
The Google Vertex API is widely adopted to generate multilingual captions for images.
The API outputs reflect established industry standards, having gone through extensive quality checks.

\subsection{Semantic variation across languages}
\label{subsec:semvar-lang}

\noindent \textbf{Method:} We apply the same method as described in \S\ref{subsec:data-semvar} to measure semantic variation, but substituting XM captions for LLaVA and Vertex-generated captions.

\begin{figure*}[t]
    \centering
    \includegraphics[width=0.75\textwidth]{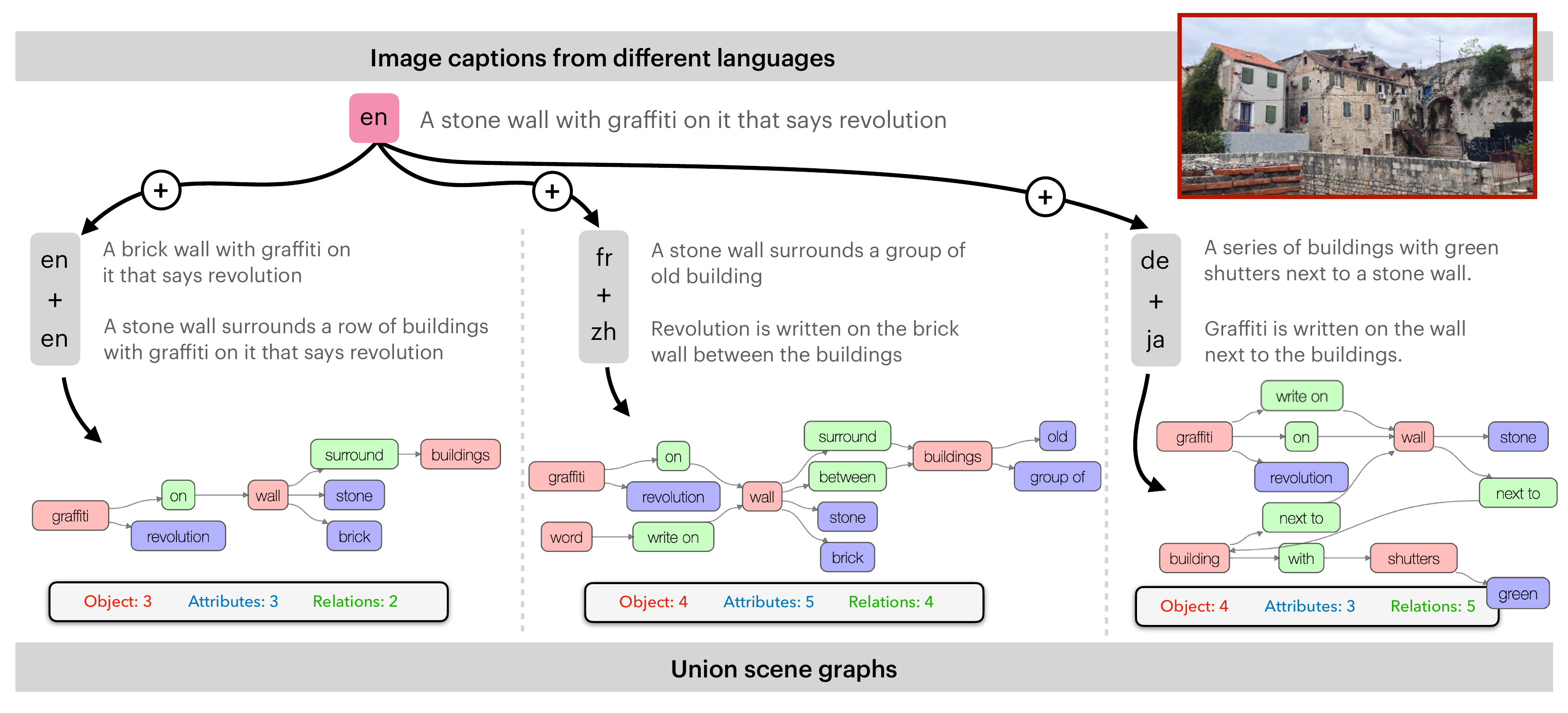}
    \caption{\textit{Semantic content evaluation.} Captions of an image in different languages and their scene graphs, when unioned together produce more varied and complex scene graphs for multilingual distributions than monolingual ones. Captions from Vertex.}
    \label{fig:scene-graph_union}
\end{figure*}

\noindent \textbf{Results:}
We find that \textbf{multilingual unioned scene graphs cover more objects, relations, and attributes than monolingual ones}~(Table~\ref{tab:llava_vertex_combined_evaluation}).
For instance, unioning the scene graph from an English caption with the scene graphs from a French and a Chinese caption increase the resulting scene graph by
$35.2\%$ objects, 
$28.0\%$ relations,
and $45.6\%$ attributes for LLaVA captions; and by
$13.2\%$ objects,
$16.6\%$ relations,
and $37.1\%$ attributes for Vertex API captions.
The smaller increase for Vertex API captions is likely explained by the more standardized (lower-variance) captions produced by Vertex API compared to LLaVA.
Figure~\ref{fig:scene-graph_union} displays an example of such a scene graph.

\subsection{Can multilingual variation be explained by `different language modes'?}
There may be a concern that multilingual captions generated in different languages by models are diverse because they are `(built to) operate in different modes' for each language; for example, tokens from different languages might `activate' language-specific representation axes \cite{chang2022geometry}, or APIs might route caption requests in different languages to model instances finetuned/trained on different data.
Can all of the increase in concept coverage (as measured by larger scene graphs) for multilingual captions over monolingual captions be explained \textit{solely} by models being built to work in `different modes' for different languages?
If the answer is `yes', then the increase seems uninteresting.
Although there are difficulties in measuring this, we make an argument that the answer is `no'.
To do so, we compare multilingual captions produced by one model (LLaVA or Vertex API) against English captions produced by multiple models (e.g., LLaVA, GIT\cite{Wang2022GITAG}, BLIP~\cite{li2022blip}) and find that their variation/coverage to be similar.
One would expect the variation among \textit{multilingual captions produced by one model} to be substantially smaller than \textit{monolingual captions produced by different models}, since the difference between caption variation due to \textit{`different language modes' inside one model} should be much smaller than the caption variation due to \textit{totally different training procedures, model architectures, etc. across different models} that happen to be in the same language~\citep{conneau-etal-2020-emerging, XU2023126287}.
However, if the variation between the two is \textit{roughly the same}, then we cannot explain the variation among multilingual captions produced by one model to the model operating in `different language modes'.

\noindent\textbf{Method:} Formally, let $c_{i, m}^l$ be a caption produced by model $m$ for image $i$ in language $l$.
Let $m$ either be LLaVA or Vertex.
For a general variation metric $\mathcal{M}$ (e.g. size of unioned scene graphs or expressive coverage), we want to compare $\mathbb{E}_{i \in I, L \in \mathcal{L}; |L|=3}[ \mathcal{M}(\{ c_{i, m}^l : l \in L \}) ]$ (variation metric $\mathcal{M}$ across multilingual captions produced by the same model $m$) and $\mathbb{E}_{i \in I} [\mathcal{M}(\{ c_{i, m}^{\text{en}}, c_{i, \text{GIT}}^{\text{en}}, c_{i, \text{BLIP}}^{\text{en}} \})]$ (variation metric $\mathcal{M}$ across English captions produced by different models).
If these two values are similar, then the variation in LLaVA and Vertex's multilingual outptus cannot be explained by models operating in `different language modes'.



\noindent\textbf{Results:} 
Our results suggest that demonstrated differences in scene graphs constructed from models' captions \textbf{are not entirely explained by `different language modes'}.
Interestingly, we find that the same-model across-language scene graphs are only slightly smaller than the same-language across-model scene graphs:
$92.4\%$ objects, 
$98.4\%$ relations, 
and $97.7\%$ attributes.
See Table~\ref{tab:multimodel-comparison} for more details.
This supports the significance of caption content differences across languages.
Furthermore, we find that caption sets generated by one model (Vertex) in different languages have large proportions of concepts not covered by English GIT and BLIP models:
$47.8\%$ of objects,
$71.4\%$ of relations,
and $59.5\%$ of attributes.
See Table~\ref{tab:multimodel-intersections} for more details.

\begin{table*}[t]
    \footnotesize
    \centering
    \caption{Ground truth coverage ($\vert \mathbb{C} \cap \mathbb{G} \vert / \vert \mathbb{G} \vert$) increases when sampling multilingual captions. $\mathbb{C}$ refers to the caption concept set and $\mathbb{G}$ refers to the `ground truth' Visual Genome concept set. $\vert \mathbb{C} \cap \mathbb{G} \vert$ (unnormalized intersection size, number of objects shared) is provided for reference. `avg' is the mean across \textit{all} triplets of languages.}
    \begin{tabular}{c c cccccccp{0.05cm} cccc}
    \toprule
    & \multicolumn{1}{c}{} & \multicolumn{7}{c}{\textbf{Monolingual}} && \multicolumn{4}{c}{\textbf{Multilingual}} \\
    \cmidrule{3-9}\cmidrule{11-14}
     & & $3\times$en & $3\times$de & $3\times$fr & $3\times$ru & $3\times$zh & $3\times$ja & $3\times$ko && en,fr,zh & fr,ru,zh & de,fr,ru & avg \\
    \midrule
    
    \multirow{2}{*}{\cellcolor{white}Vertex} & \cellcolor{white} $|\mathbb{C} \cap \mathbb{G}| / |\mathbb{G}|$ & 
    9.6\% & 9.7\% & 9.9\% & 10.3\% & 9.8\% & 9.9\% & 9.3\% && 11.7\% & 11.7\% & 11.2\% & 11.5\% \\
    & $|\mathbb{C} \cap \mathbb{G}|$ & 1.55 & 1.60 & 1.62 & 1.69 & 1.60 & 1.59 & 1.49 && 1.92 & 1.92 & 1.84 & 1.90 \\
    \midrule
     \multirow{2}{*}{\cellcolor{white}LLaVA} & \cellcolor{white} $|\mathbb{C} \cap \mathbb{G}| / |\mathbb{G}|$ & 
     9.5\% & 10.9\% & 11.6\% & 9.3\% & 9.1\% &-&-&& 12.0\% & 12.1\% & 12.5\% & 12.0\% \\
    & $|\mathbb{C} \cap \mathbb{G}|$ & 2.86 & 3.30 & 3.55 & 2.83 & 2.75 &-&-&&  3.71 & 3.70 & 3.86 & 3.69 \\
    \bottomrule
\end{tabular}
\label{visual-genome}
\end{table*}

\subsection{Grounding variation in scene graphs}

Another concern may be that captions generated by a model might contain concepts that are not `objectively' grounded in the image.
For example, one caption might mention a `tree', another might mention `tree trunk' and `branches', possibly to describe specific attributes of each (e.g., the `tree trunk is brown, and the branches are black') or describe a unique relation between them (e.g., `the branches are all on the same side of the tree trunk').
Although these types of differences are important, since they reflect complementary information about the content of the image, if variation across caption content and expression can be explained \textit{solely} by variation induced by these kinds of `subjective' focuses, then one could argue it is not as significant or interesting.
Therefore, we reproduce our scene graph experiment to measure content variation \textit{only on object sets documented in the Visual Genome object list}~\citep{Krishna2016VisualGC} for that image.
Since the Visual Genome object list is standard and shared across all models, it provides an `objective' grounding for measuring the size of scene graphs by objects.

\noindent\textbf{Method:} Formally, let $VG_i$ be the Visual Genome object set for an image $i$. 
Recall from \S\ref{subsec:semvar-lang} that $\text{mono}_i^l$ and $\text{multi}_i^L$ are the monolingual and multilingual caption sets using language $l$ and language set $L$, respectively, for image $i$.
Let $\textsf{Obj}(C)$ take in a caption set and output the unique objects (as if constructing a unioned scene graph and then extracting only the objects).
We want to compare
$\mathbb{E}_{i \in I, l \in \mathcal{L}}[|\textsf{Obj}(\text{mono}_i^l \cap VG_i)|]$ ($\#$ unique VG objects covered by monolingual caption set) and
$\mathbb{E}_{i \in I, L \in \mathcal{L}}[|\textsf{Obj}(\text{multi}_i^L \cap VG_i)|]$ ($\#$ unique VG objects covered by multilingual caption set).
If the latter is significantly larger than the former, then this is a stronger confirmation that there tend to be distributional non-overlaps/differences across concepts covered in captions across languages.




\noindent\textbf{Results:} The results demonstrate, like before, that \textbf{multilingual scene graphs capture more Visual Genome objects/concepts than monolingual scene graphs} (see Table~\ref{visual-genome}).
For example, unioning an English scene graph with a French and Chinese scene graph captures
$23.9\%$ more Visual Genome-annotated objects than unioning with two other English scene graphs.
Interestingly, we also find some evidence that multilingual scene graphs are able to identify objects which are unidentified even in Visual Genome's dense annotations~(App. Table~\ref{tab:vg-exceed-examples}).

\subsection{Expressive variation across languages}
\label{subsec:expvar-lang}

\noindent\textbf{Method:} We apply the same method as described in \S\ref{subsec:data-exprvar} to measure semantic variation, but substituting XM captions for LLaVA and Vertex-generated captions.

\noindent\textbf{Results:} Our results demonstrate \textbf{variance in models' caption expression across languages}. 
We find that across all metrics, expressive coverage is generally higher for multilingual rather than monolingual subsets (Table~\ref{tab:llava_vertex_combined_evaluation}).
For example, the range of noun concreteness in caption sets widens by $8.9\%$ for multilingual sets on average for LLaVA, and by $13.1\%$ for Vertex.
We also find that the mean width of caption sets in model representations is larger for multilingual sets than monolingual sets.
For example, the average coverage in embedding space of English LLaVA captions increases from 0.22 to 0.45 when switching to an English, French, and Chinese caption set, and from 0.19 to 0.37 for Vertex.
\section{Effects of Fine-tuning Across Language}
\label{finetuning}

In \S\ref{sec:dataset-variation} and \S\ref{sec:model-var}, we showed that dataset and model captions exhibit semantic and expressive variation across languages.
What effect does this have on how we train and fine-tune captioning models? Do models internalize the specific distributional characteristics of captions from one language?

\noindent\textbf{Method:} To evaluate this, we finetune a pretrained vision-language model, GIT~\cite{Wang2022GITAG} -- chosen for its simple architecture and robust performance across benchmarks, on 8 constructed image caption sets. 
These finetuning datasets use the same set of 1.8k training images from the Crossmodal dataset (``XM'')~\cite{Thapliyal2020CrossmodalLG}.
The first 7 training sets contain translations from a single language (e.g.~1.8k captions all from French, 1.8k captions all from Chinese).
The eighth dataset is constructed with equal proportions of captions across all 7 languages.
We inference the finetuned models on a hold-out set of 1.8k images from XM and evaluate performance against each of the the corresponding 8 sets of image caption labels.
We use SPICE F1-score~\cite{Anderson2016SPICESP}, which measures conceptual overlap over low-level syntax adherence, as our evaluation metric.
Validating each of the 8 fine-tuned models on each of the 8 holdout caption sets yields 64 evaluations.
We perform this entire process on captions from XM, LLaVA, and Vertex.

We interpret our results using the following logic.
Let $\mathcal{S}(A; B)$ represent the score of a model fine-tuned on the data from language $B$ on the evaluation data from language $A$ (suppose bigger is better).
If we observe superior model test performance on the test set corresponding to the language it was trained on relative to other test sets (that is, $\forall_{A \in \mathcal{L}, B \in \mathcal{L}; A \neq B} \mathcal{S}(A;B) \ll \mathcal{S}(A; A)$), the model has internalized distributional characteristics of each language.

\noindent\textbf{Results:} We find that a \textbf{model fine-tuned on language $X$ performs best on language $X$} (Table~\ref{tab:distillation_results_vertex}) for Vertex captions.
We find the same results across models finetuned on data from 
LLaVA (Table~\ref{tab:finetuning_llava}), and from the human annotated XM dataset (Table~\ref{tab:finetuning-xm}), including when evaluating with default loss rather than SPICE F-score.
Moreover, finetuning on a multilingual collection of captions yields a model which performs consistently well across all evaluation data compositions.
In the case of training on human-generated multilingual captions (from the Crossmodal dataset), this model is the \textit{second-best} model for 5/7 holdout-set languages out of 8 models, coming only behind a model finetuned on that language itself. 
Together, these suggest that models trained with multilingual data may represent diverse perceptual modes and learn more about the visual world.

\begin{table}[!t]
    \scriptsize 
    \setlength{\tabcolsep}{3pt} 
    \centering
    \caption{\textbf{Model outputs evaluation results.} SPICE F-scores when evaluating a model fine-tuned on the training set from the language on the left against the validation set from the language on the top. `multi' refers to an even split across all languages. \first{Red}: best performance on a split, \second{yellow} highlights model fine-tuned on `multi'.}
    \def\arraystretch{1.15}
    \begin{tabular}{cr cccccccc}
        \toprule
     & \multicolumn{1}{c}{} & \multicolumn{8}{c}{\textbf{Evaluated on}}
        \\
         & & \textbf{en} & \textbf{de} & \textbf{fr} & \textbf{ru} & \textbf{zh} & \textbf{ja} & \textbf{ko} & \textbf{multi} \\[0.5mm]
        & \textbf{en} & \first{0.271} & 0.225 & 0.229 & 0.219 & 0.218 & 0.229 & 0.232 & 0.230 \\
        & \textbf{de} & 0.213 & \first{0.245} & 0.219 & 0.217 & 0.215 & 0.210 & 0.226 & 0.219 \\
        & \textbf{fr} & 0.248 & 0.240 & \first{0.259} & 0.234 & 0.236 & 0.239 & 0.253 & 0.246 \\
        & \textbf{ru} & 0.226 & 0.234 & 0.228 & \first{0.254} & 0.231 & 0.236 & 0.237 & 0.239 \\
        & \textbf{zh} & 0.199 & 0.202 & 0.199 & 0.207 & \first{0.247} & 0.220 & 0.224 & 0.216 \\
        \rot{\rlap{\textbf{~Fine-tuned on}}}
        & \textbf{ja} & 0.212 & 0.212 & 0.215 & 0.212 & 0.226 & \first{0.266} & 0.245 & 0.223 \\
        
        & \textbf{ko} & 0.218 & 0.222 & 0.224 & 0.217 & 0.242 & 0.239 & \first{0.271} & 0.235 \\
        & \textbf{multi} & \second{0.239} & \second{0.233} & \second{0.234} & \second{0.233} & \second{0.235} & \second{0.243} & \second{0.252} & \second{0.244} \\
        \bottomrule
    \end{tabular}
    \label{tab:distillation_results_vertex}
\end{table}
\section{Discussion}


\subsection{Implications}

\noindent\textbf{Reframing the ``curse of multilinguality''.}
We contribute another framing to the ongoing debate related to the ``curse of multilinguality.''
On one hand, arguments in favor of multilingual language and vision models are often made from an accessibility perspective of expanding information and technology access to individuals~\cite{Hu2020XTREMEAM}.
Meanwhile, empirical evidence suggests that increasing the number of languages a model handles degrades monolingual performance~\cite{Conneau2019UnsupervisedCR}.
Multilingualism is, seen as a trade-off against model performance.
Our work provides support for the possibility of a system benefit from including multilingual data. 
Using ``organically'' / human-produced multilingual data introduces substantive changes in the semantic concepts mentioned for the same images. 
Thus, incorporating the diverse content distribution may improve model robustness~\cite{kariyappa2019improving,ramanujan2023connection,taori2020measuring,nguyen2024multilingualdiversityimprovesvisionlanguage}.
$\circ$
\textbf{Translating English content to other languages may be missing some information.}
Many efforts to build vision-language models for a non-English language $X$ train on translations from English to $X$~\cite{Mishra2021AHI,Chen2022PaLIAJ,Geigle2023BabelImageNetMM,Changpinyo2022TowardsMV,Pfeiffer2021xGQACV,Qiu2022MultilingualML}.
Although these systems are usable in $X$, our investigation suggests that they may miss out on the unique characteristics of an ``organically $X$'' content distribution (i.e., that native $X$-speakers would provide).
Indeed, other works which collect captions from $X$-speakers and train models on them suggest quantitative and qualitative improvements over training from English-to-$X$ translated captions~\cite{yoshikawa-etal-2017-stair, Wu_2019}. 
Our work provides an explanation for such observations.
$\circ$
\textbf{Wariness of a ``perceptual monoculture''.}
The English-dominated language modeling landscape may not only introduce accessibility issues for non-English speakers but also possibly a ``perceptual monoculture''~\cite{Kleinberg2021AlgorithmicMA}.
In these ways, our contribution shares motivation and aims with a diverse array of previous works in how models and datasets encode and internalize particular limiting and possibly harmful knowledge, such as by gender~\cite{Caliskan2022GenderBI}, nationality~\cite{Santy2023NLPositionalityCD}, 
and language~\cite{Wendler2024DoLW, AlKhamissi2024InvestigatingCA}.



\subsection{Recommendations }



\noindent \textbf{Multilingual modeling from organically multilingual data.} 
Currently, many multilingual vision-language models are built by training models on captions translated from English into the desired language. 
When possible, multilingual vision-language applications should be built by training models collected from native speakers. 
Our results suggest that the advantage is not only \textit{syntactic} (i.e. more ``natural-sounding'' than translations), but that annotations produced in the desired language have specific distributional \textit{semantic} and \textit{expressive} characteristics.

\noindent  \textbf{Multilinguality as visual descriptive diversity.} 
Multilingual vision-language datasets should be explored as repositories of perceptually diverse annotations. 
At a basic level, simply training models on multilingual rather than monolingual vision datasets may expose them to a more diverse, rich set of concepts. 
At a more complex level, specifically designed structured learning paradigms such as contrastive learning may be able to exploit concept diversity across multilingual captions to yield richer image representations.


\subsection{Conclusion}

This paper demonstrates semantic and expressive in image captions across languages, revealing meaningful differences in how visual content is represented.
Our results support that training models on multilingual captions exposes them to a wider range of semantic and expressive material.
This increased range may result in more robust vision representations \cite{nguyen2024multilingualdiversityimprovesvisionlanguage}.
By embracing the richness that different languages bring, we can move toward more inclusive, nuanced models that better reflect the diversity of human perception.


\section*{Acknowledgements}

We thank Katharina Reinecke for the many insightful discussions that inspired this work; Quan Ze (Jim) Chen, Matthew Wallingford, Sarah Pratt, Thao Nguyen, Hyunwoo Kim, Wei-Chiu Ma, Chris Lin, Lisa Orii, Nino Migineishvili, Ashish Sharma, Jae Sung (James) Park and members of RAIVN lab for the multiple rounds of discussion, feedback and reviewing; Cheng-Yu Hsieh and Ashish Sharma for code/package assistance; Allen AI for running the surveys; and Chun-Liang Li at Google Cloud for running the Vertex API. We also thank participants of our study for carefully captioning the images.
{
    \small
    \bibliographystyle{ieeenat_fullname}
    \bibliography{main}
}

\clearpage
\setcounter{page}{1}
\maketitlesupplementary

\section{Related work}
\label{related-work}

Existing work in \textbf{multilingual multimodal modeling} investigates how vision-language models can perform better across a variety of languages.
Many previous works have proposed methods to build non-English and multilingual models for specific vision tasks such as captioning, question answering, and retrieval~\citep{Hitschler_2016, gao2022unisonunpairedcrosslingualimage, 9223087, Emami2022Arabic, chen2022altclipalteringlanguageencoder}.
To benchmark and build more multilingual models, many multilingual vision datasets have been introduced~\citep{li2019cococncrosslingualimagetagging, Geigle2023BabelImageNetMM, Luu_Thuy_Nguyen_2023, ku2020roomacrossroommultilingualvisionandlanguagenavigation, atuhurra2024constructingmultilingualvisualtextdatasets}.
Many more recent large vision models are trained to be multilingual~\citep{Chen2022PaLIAJ, ilharco_gabriel_2021_5143773, geigle2024mblipefficientbootstrappingmultilingual}.
These models have been probed for biases across language and associated cultures~\citep{ananthram2024perspectivediagnosingwesterncultural, espana-bonet-barron-cedeno-2022-undesired, Hinck2024WhyDL}. 
To better measure and counteract these biases, vision datasets have been built which include images captured from diverse geographical regions around the world~\cite{Atwood2019TheII, Schumann2021AST,Liu2024SCoFTSF,Liu2023TowardsER}, and which create diverse visual knowledge by annotating images with culture- and region-specific information, such as identifying regional dishes, dresses, and ideas~\cite{Fu2022TheresAT,liu2021visually, mohamed2022artelingo,Yin2023GIVLIG}.
We build upon this rich lineage of multilingual vision work: rather than seeking to \textit{propose new multilingual datasets} (which may offer new concepts) or \textit{expand vision models' capabilities on non-English languages}, we seek to demonstrate that \textit{multilingual vision datasets and models may already exhibit meaningful information differences across languages}.

Our inquiry is inspired by \textbf{research in cross-cultural Psychology}.
Psychologists, anthropologists, and philosophers provide strong evidence that salient visual features differ systematically across cultures and languages, with broadly ranging studies including cross-cultural psychology~\cite{Nisbett2003CultureAP,Koo2018AnalyticVH}, usage-based linguistics~\cite{tomasello2006usage,langacker2019construal}, organizational principles of perception and cognition~\cite{wertheimer1912experimentelle,kohler1967gestalt}, and the cognitive realities of one's perceptual experience~\cite{Wittgenstein1953-WITPI-4, hill2022perceptual, talmy2018fictive, husserl2012cartesian}. 
Cognitive linguistics posits direct ties between what we say to the way we think and perceive the world~\cite{boroditsky2006linguistic,lakoff2008metaphors,lakoff1987fire}. 
They further suggest that expressed meaning depends on not only \textit{what} is said --- the \textit{semantic content} of what we say, but also equally on \textit{how} we say it --- the very \textit{manner of expression} we choose to say it (e.g., specificity of word choice, tone and mood of expression)~\cite{langacker2019construal, trott-etal-2020-construing, boroditsky2003sex}. 
That is, a speaker's conceptualizations has direct influence over what linguistic features or words they reach for and how put them together when they formulate our thoughts into words~\cite{langacker1993universals,TALMY198849,croft2000construal}.
This leads us to study both \textit{semantic} and \textit{expressive} variation of captions across languages, e.g. in \S\ref{subsec:data-semvar} and \S\ref{subsec:data-exprvar}.

In general, \textbf{human-centric approaches} to computer vision center around considerations of human abilities and limitations in the development of models and applications.
For example, methods highlighting saliency attempt to identify which image regions and features people find most important~\cite{Spain2008SomeOA,Elazary2008InterestingOA,Berg2012UnderstandingAP,Yun2013StudyingRB,Tavakoli2017PayingAT}.
User-centric vision modeling adapts the models to user-specific preferences and knowledge~\cite{Stretcu2023AgileMF, Cohen2022ThisIM,toubal2023modeling,Ratner2017SnorkelRT}.
Similarly, our work looks closely at the differences between populations of humans ``behind'' multilingual vision datasets (and downstream multilingual vision models).

\section{Limitations}
\label{limitations}


\noindent \textbf{Our chosen 7 languages.} Our selection of languages is diverse but not representative of global linguistic diversity. 

\noindent \textbf{Mid-sized scale.} Our experiments operate at a mid-sized scale (thousands of images), emphasizing breadth in languages over depth in images. 
Future studies may forego such a wide exploration to investigate more specific phenomena at a larger image scale, such as if models differ in their image understanding when trained on captions from different languages. Previous works have shown promise in this direction by showing how better-quality, denser, and more diverse captions can help with better image understanding~\cite{nguyen2023improving,lai2023scarcity}.

\noindent \textbf{Risk of linguistic essentialism.} Categorizing differences solely with languages may pose a risk of essentializing or stereotyping them, suggesting that all members that speak a language describe the world similarly. 
We emphasize that we do not make \textit{categorical} but rather \textit{distributional} claims, aiming to show general differences across a large set of samples.

\section{Experimental Details}

\subsection{Translation into English}
\label{translation_quality_check}

We prompted GPT-4~\cite{OpenAI2023GPT4TR} to translate text with: ``Return the translation (and only the translation) of the following text from \verb|[SRC_LANG]| into \verb|[TGT_LANG]| exactly with all details: \verb|[TEXT]|''.
We find that this prompt produces translations which especially preserve the conceptual details of the original text.

Although some language-specific meanings will inevitably be lost in any translation between languages, we ensure that our English translations are as faithful as possible to the concepts expressed in the original language by conducting a human evaluation.
We recruit 2-3 speakers for each of the six non-English languages (French, German, Russian, Chinese, Japanese, Korean), fluent in both the original image and English.
Each subject evaluates 30 pairs of original and translated text. Of these 30 pairs, 10 are Vertex captions on Crossmodal images, 10 are LLaVA captions on Crossmodal images, and 10 are Vertex captions on Visual Genome images. This composition ensures wide coverage across image domains and caption format.
Each translation evaluation has two parts. 
Firstly, subjects annotate the overall translation quality on a 1 to 5 scale, in which 1 is ``entirely inaccurate'', 2 is ``some of the information is preserved'', 3 is ``only the most important information is preserved'', 4 is ``most of the information is preserved (the translation is adequate but not perfect)'', and 5 is ``entirely accurate''.
Secondly, subjects examine 11 general categories of concepts in natural visual scenes, provided by TIFA~\cite{Hu2023TIFAAA}: objects, animals/humans, attributes, activities, spatial relations, counting, food, materials, shapes, locations and colors.
Subjects mark each category either as ``Good'' (the concept was present in the original text and faithfully represented in the translation), ``Missed'' (the concept was present in the original text but absent or not faithfully represented in the translation), or ``N/A'' (the concept was not present in the original text).
Table \ref{tab:translation_results} demonstrates that the translations are nearly entirely accurate, especially for European languages, and preserve nearly all of the salient content categories for understanding visual scenes.

Annotators were allowed to provide free-text explanations for areas in which the translation was inadequate. 
We provide a random sampling of comments to provide a holistic idea of the translation weaknesses.
Overall, the changes to the translations indicated in the comments do not change the content or expression of the text in a substantive way.

One possible confounder in results like \S\ref{finetuning} is that language-specific syntactic artifacts introduced during translation.
For instance, text translated from German into English might have a unique syntactic structure which distinguishes it from text originally written in English.
If this is the case, then it should be possible to identify translated text from one language versus another.
To test this limitation, we embed all translated captions using a BERT-based model~\cite{reimers-2019-sentence-bert}.
We fit a logistic regression model to predict a sample's original language from these features, and find near-random chance performance at $16.43\%$ (random chance is $1/7 \approx 14.29\%$).
This suggests that the translation artifact confounder does not explain the observed results.

\subsection{Probing Multilingual Capabilities in LLaVA}
\label{app:llava-multilingual}

Models like LLaVA which are trained/fine-tuned with English data but which include multilingual LLM components can retain some of these multilingual capabilities.
In order to request LLaVA generate captions in a target language, we change the prompt at all levels to correspond to that language language, displayed in Table~\ref{tab:llava-prompting}.
This works successfully across each of the non-English languages considered in this work, except for Korean and Japanese, which exhibit significantly worse quality.

\subsection{Image Captioning User Study}
\label{image-captioning-user-study}
\begin{figure*}[!h]
    \centering
    \begin{subfigure}{.45\textwidth}
      \centering
      \includegraphics[width=6cm]{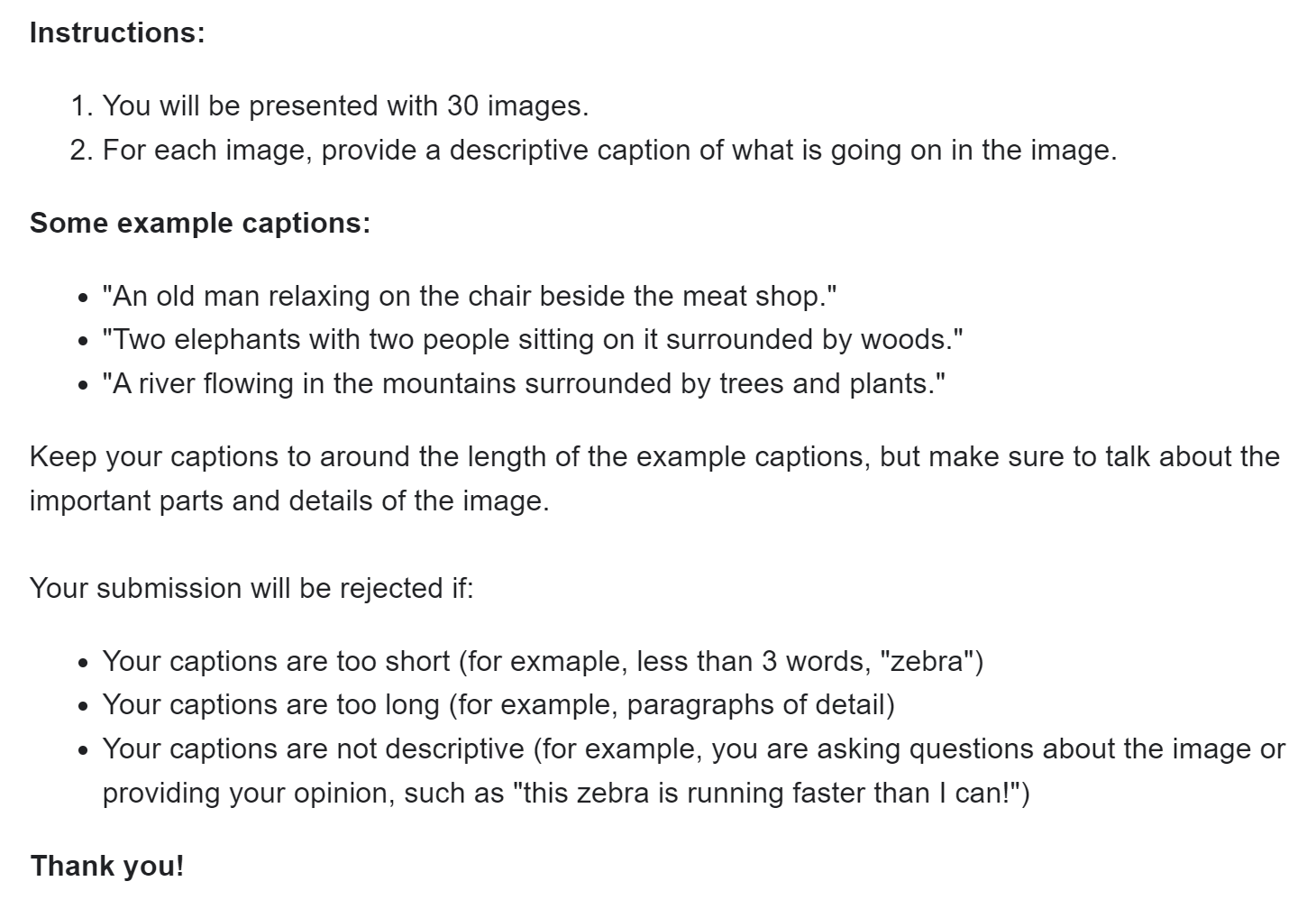}
      \caption{English instructions.}
      \label{fig:instructions}
    \end{subfigure}%
    \hfill
    \begin{subfigure}{.5\textwidth}
      \centering
      \includegraphics[width=6cm]{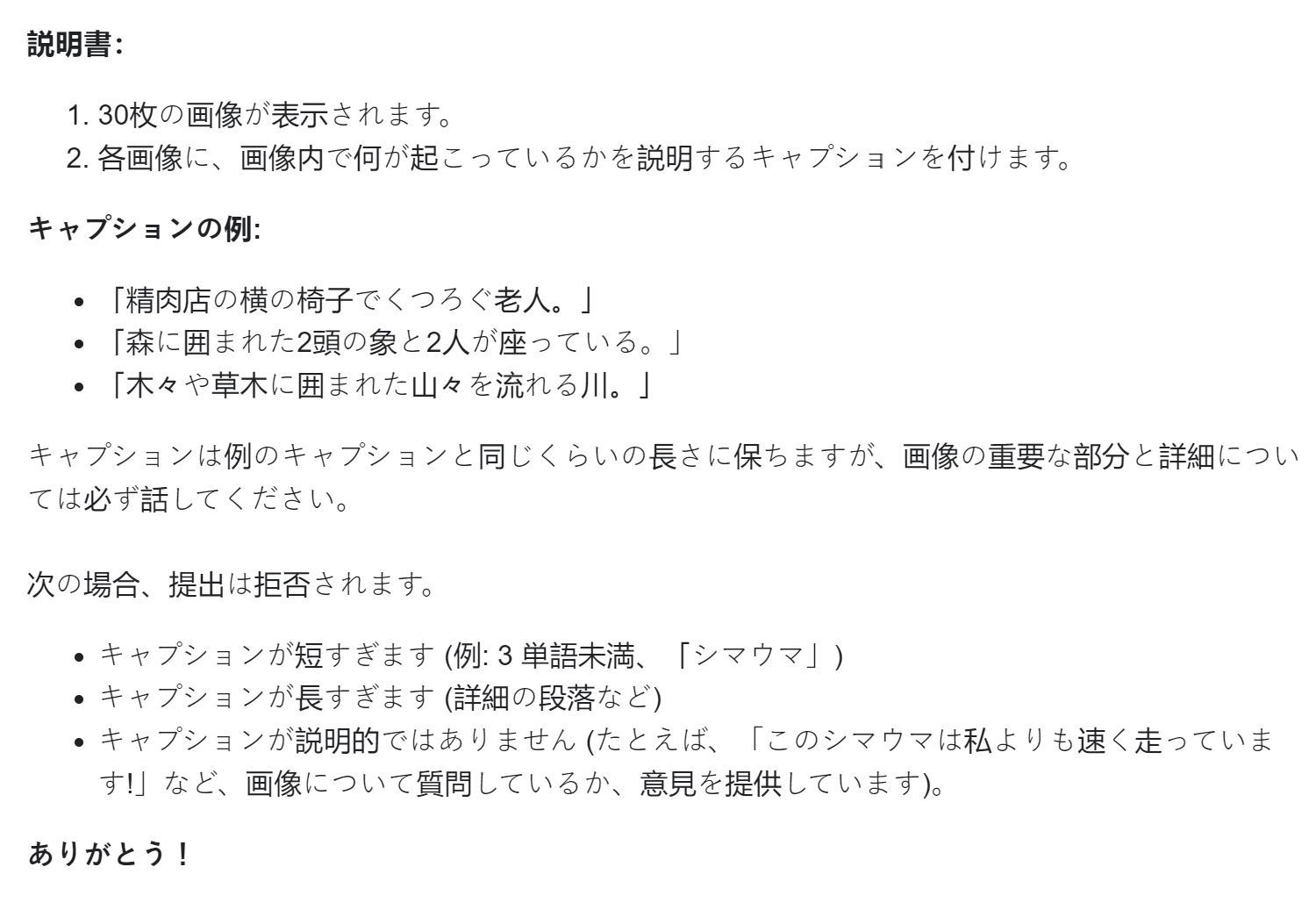}
      \caption{Japanese instructions.}
      \label{fig:instructions2}
    \end{subfigure}
    \caption{Instructions and examples presented to human evaluation participants for image captioning.}
\end{figure*}

We recruited 10 English speakers from the US and 10 Japanese speakers from Japan. 
The instructions given to them are presented in Figures~\ref{fig:instructions} and \ref{fig:instructions2}.
A sample of the produced captions is given in Table~\ref{fig:user-study-examples}.

Because large-scale image-text datasets do not conduct much annotator information, it is difficult to make detailed and strong inferences about the psychological causes of the observed results, so more work is needed in this direction.
However, as a start, we recruited 10 English speakers from the U.S. and 10 Japanese speakers from Japan to caption 30 Visual Genome images and repeated the semantic content evaluation for human-produced captions.
We find, in the same pattern as before with model captions, that unioning English scene graphs with Japanese scene graphs expands the size by
$8.4\%$ objects, 
$7.7\%$ relations, 
and $6.5\%$ attributes 
over unioning English scene graphs with other English scene graphs.
Moreover, a manual inspection of the captions suggests that the captions roughly echo the predictions from cross-cultural perceptual psychology -- Japanese captions tend to mention background objects and information more than English ones~(see Figures~\ref{fig:intro-fig} and \ref{fig:user-study-examples}).

\section{Supplementary Data and Figures}

Our results across all evaluations are displayed in Table~\ref{primary-table}.

\subsection{Semantics Evaluations}
\label{app:semantics}

\begin{figure*}[p]
\begin{subfigure}{.32\linewidth}
\centering
\includegraphics[width=\textwidth]{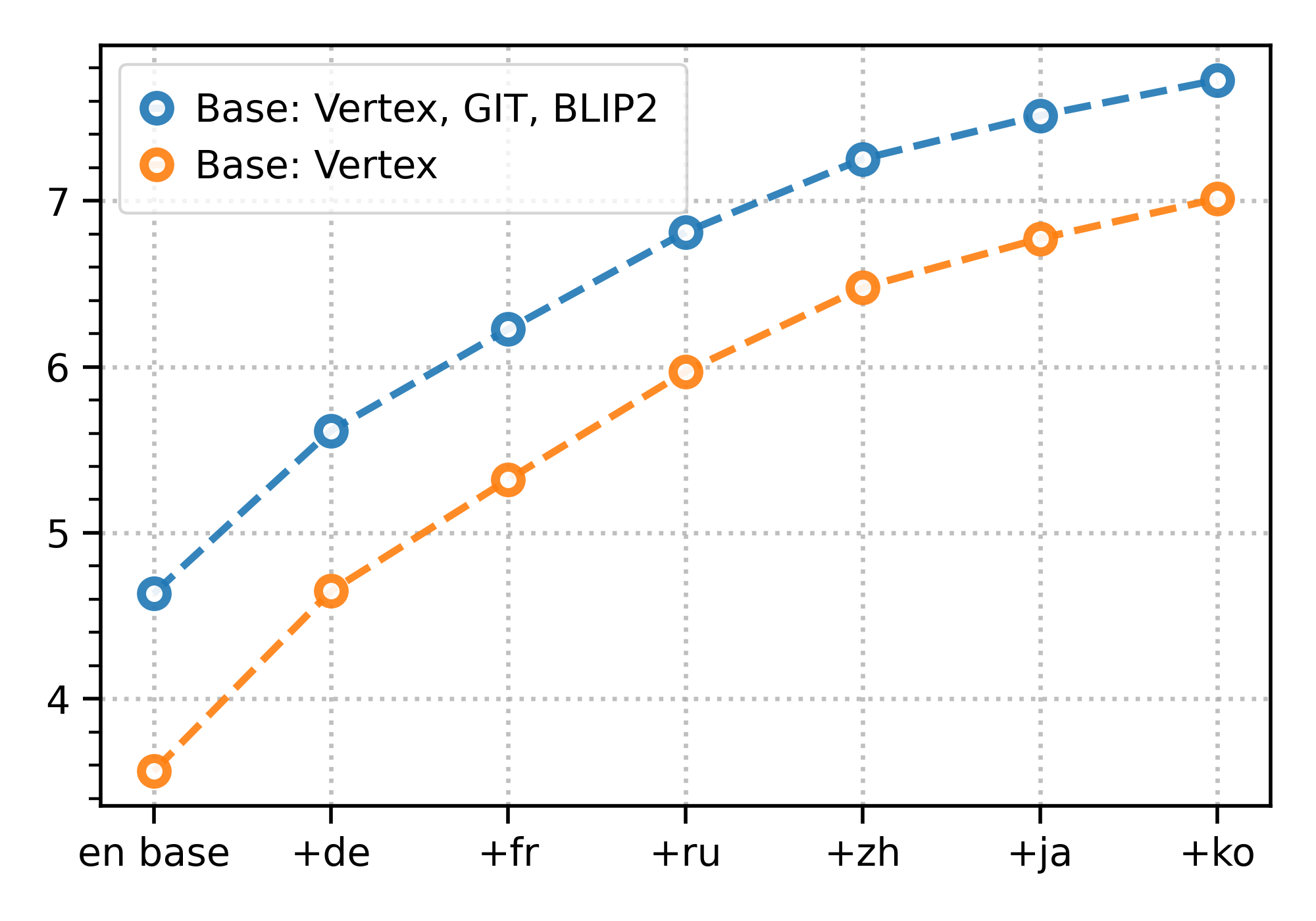}
\caption{\# objects in scene graph}
\end{subfigure}%
\hfill
\begin{subfigure}{.32\linewidth}
\centering
\includegraphics[width=\textwidth]{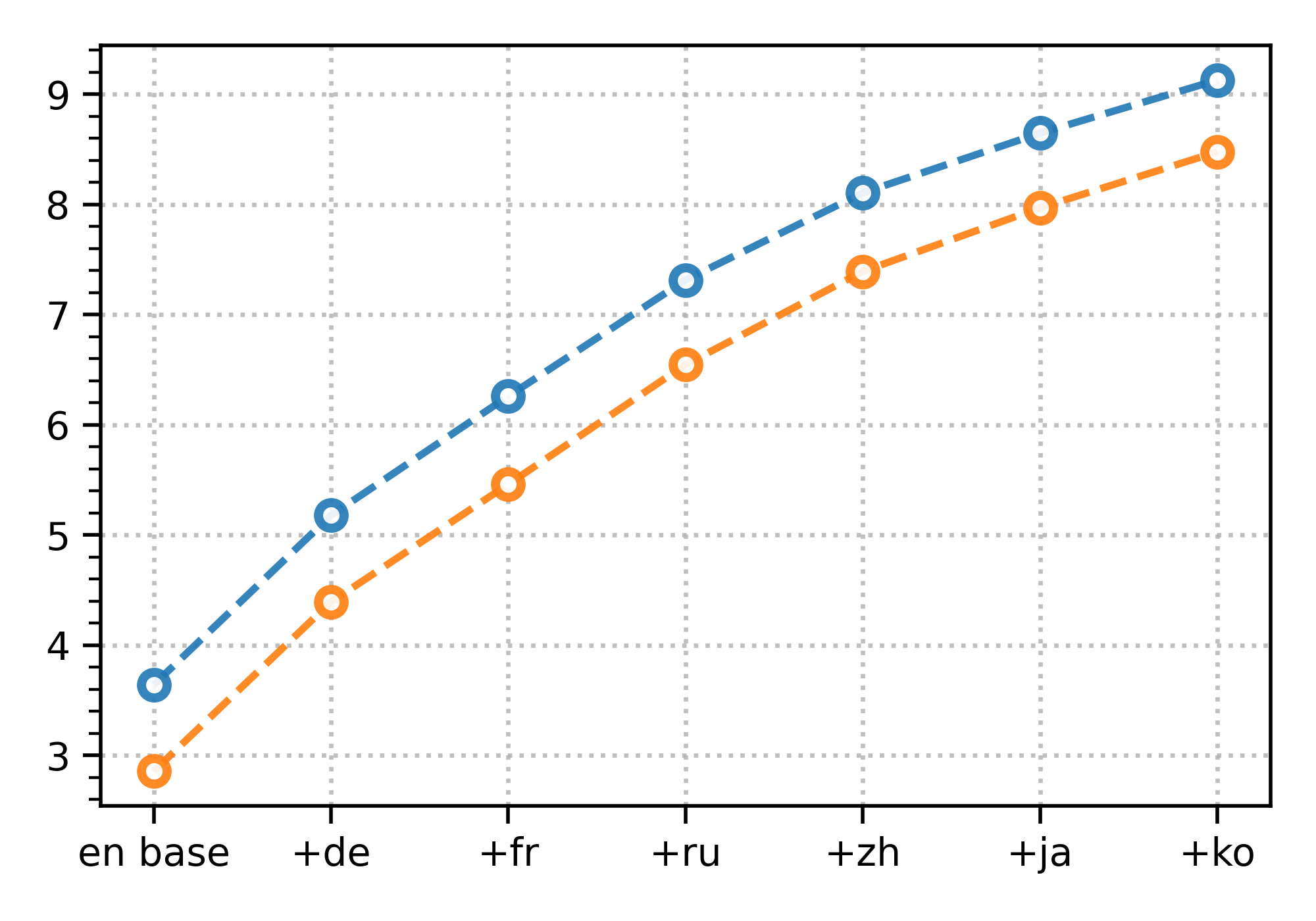}
\caption{\# relations in scene graph}
\end{subfigure}
\hfill
\begin{subfigure}{.32\linewidth}
\centering
\includegraphics[width=\textwidth]{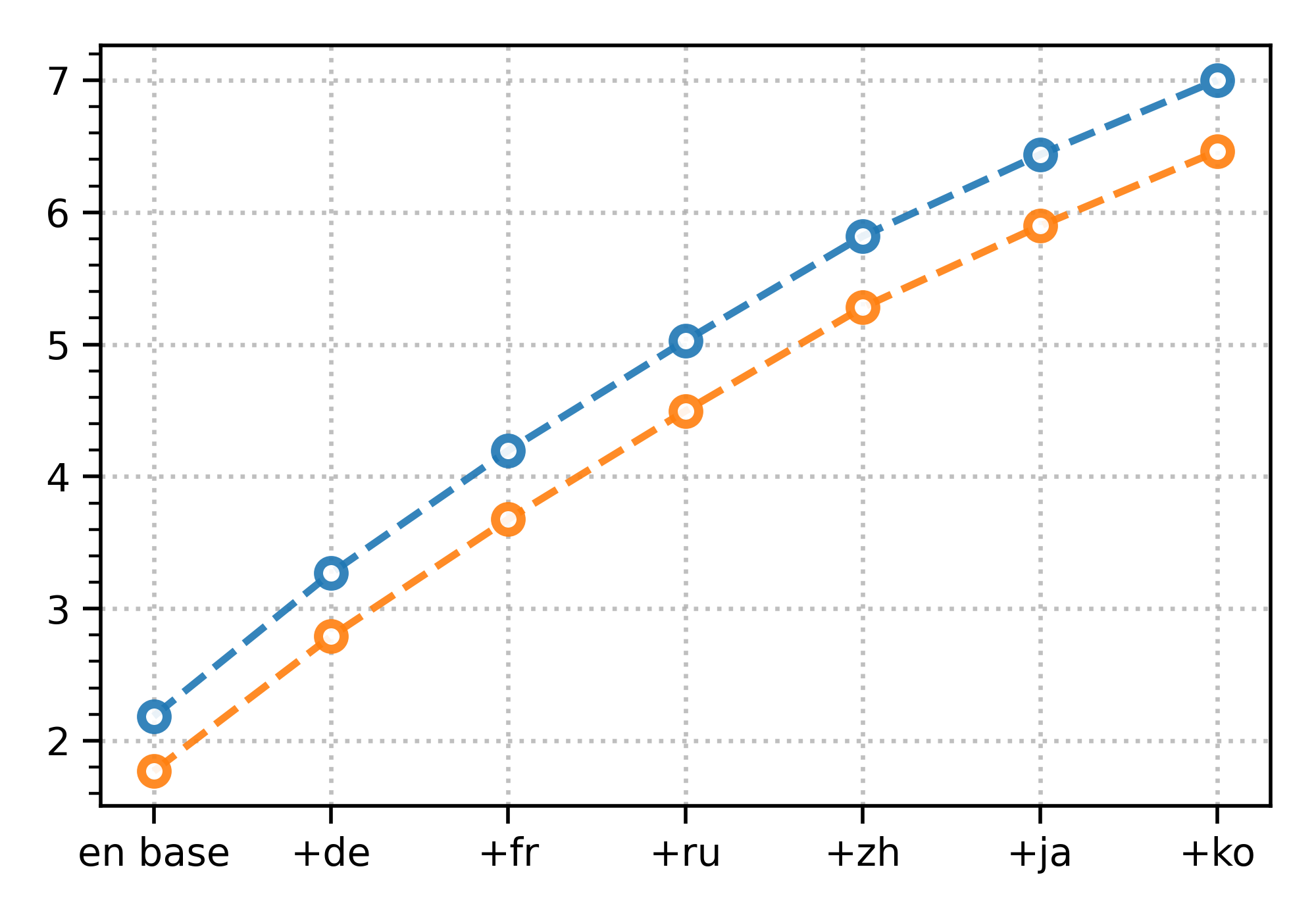}
\caption{\# attributes in scene graph}
\end{subfigure}\\[1ex]
\vspace{-5mm}
\caption{Scene graphs of captions unioned cumulatively from different languages lead to more coverage in objects, relations, and attributes.}
\label{fig:progressive_growth_unioning}
\end{figure*}

Figure~\ref{fig:progressive_growth_unioning} shows that despite an expected diminishing-returns trajectory, continuously unioning even a well-developed existing scene graph with a new language's scene graph expands it. This suggests that different languages continue to have new information to add to the existing scene graph of visual knowledge.
Table~\ref{tab:crossling_metrics} displays the sizes of intersections between monolingual scene graphs as measured by the number of objects and relations, using the formula $M(A) + M(B) - M(A \cup B) = M(A \cap B)$. It is an alternative way to understand the conceptual overlap of different languages.
Table~\ref{tab:multimodel-comparison} shows that scene graphs constructed from captions \textit{from the same model but different languages} are only slightly smaller than those constructed from captions \textit{from the same language but different models}.
Table~\ref{tab:multimodel-intersections} shows the intersection sizes between monolingual and multi-model scene graphs.
Table~\ref{tab:vg-exceed-examples} shows some samples in which multilingual scene graphs identify objects in the image which are not mentioned in the Visual Genome annotations.
Figure~\ref{fig:big-scene-graph-spread} shows several examples of scene graphs generated in different languages for different samples.


\subsection{Multilingual Embedding Space Coverage}
\label{multilingual_embedding}

Recall from \S\ref{translation_fair_comparison} that many of the tools we use to measure semantics and expressions are not available in different languages (e.g., scene graph parsers, linguistic measures).
However, in the case of embedding space coverage, we can use multilingual embeddings rather than monolingual (English) embeddings (with translation of all captions into English).
We reproduce the expressive variation experiment described in \S\ref{subsec:data-exprvar} using multilingual embeddings \textit{without translation}, and find that the same result holds as in the main paper using English embeddings with translation~\ref{supp_multiembd}.
This provides further empirical support that translation bias does not interfere with our results.
However, note that multilingual embeddings have documented language biases~\citep{otmakhova-etal-2022-cross, chang2022geometry}, which is why we prefer to use monolingual embeddings with translation for a fairer comparison.

\begin{table}[!h]
    \centering
    \begin{tabular}{c|ccc|ccc}
        & \multicolumn{3}{c}{mono} & \multicolumn{3}{c}{multi} \\
        & en & fr & avg & en,fr,de & en,ru,zh & avg \\
        \hline
        XM & .274 & .279 & .280 & .327 & .328 & .340 \\
        LLaVA  & .475 & .507 & .521 & .704 & .795 & .753 \\
        Vertex & .340 & .321 & .321 & .600 & .647 & .612 \\
    \end{tabular}
    \label{results-table}
    \caption{Model representations experiment from the paper, repeated using multilingual Sentence-BERT without translation. `avg' is the mean cosine distance across all monolingual and multilingual caption sets; the difference is significant $(p < 0.001)$.}
    \label{supp_multiembd}
\end{table}

\subsection{Model outputs evaluations}
\label{app:outputs}

Tables \ref{tab:finetuning_llava} and \ref{tab:finetuning-xm} repeat the same fine-tuning experiment as outlined in \S\ref{finetuning}, but training on LLaVA and XM captions instead of Vertex captions.


\clearpage

\begin{table*}[!h]
\scriptsize
    \centering
    \caption{Human evaluations for translation quality using GPT-4 on multilingual captions. TIFA categories represent the mean proportion of non-N/A responses which are marked ``Good'' (as opposed to ``Missed'').}
    \begin{tabular}{clcccccc}
        \toprule
        \multicolumn{2}{c}{\textbf{Metric}} & \textbf{de} & \textbf{fr} & \textbf{ru} & \textbf{zh} & \textbf{ja} & \textbf{ko}  \\
        \midrule
        \multirow{3}{*}{Quality Ratings} & Mean & 4.95 & 4.76 & 4.82 &  4.63 & 4.48 & 4.48 \\
        & Median & 5.00 & 5.00 & 5.00 & 5.00 & 5.00 & 5.00 \\
        & 25th Percentile & 5.00 & 5.00 & 5.00 & 4.00 & 4.00 & 4.00 \\
        \midrule
        \multirow{11}{*}{TIFA Categories} & Objects & 1.00 & 0.99 & 1.0 & 0.97 & 0.98 & 0.90 \\
        & Animals/Humans & 1.00 & 1.00 & 1.00 & 1.00 & 1.00 & 1.00 \\
        & Attributes & 1.00 & 0.89 & 1.00 & 0.93 & 1.00 & 1.00 \\
        & Activities & 1.00 & 1.00 & 1.00 & 0.98 & 0.91 & 0.96 \\
        & Spatial Relations & 1.00 & 1.00 & 1.00 & 0.92 & 0.94 & 0.96  \\
        & Counting & 1.00 & 1.00 & 1.00  & 0.99 & 1.00 & 0.90 \\
        & Food & 1.00 & 1.00 & 1.00 & 1.00 & 1.00 & 1.00 \\
        & Material & 1.00 & 1.00 & 1.00 & 1.00 & 1.00 & 1.00 \\
        & Shape & 1.00 & 1.00 & 1.00 & 1.00 & 1.00 & 1.00 \\
        & Location & 1.00 & 0.98 & 1.00 & 0.98 & 0.96 & 0.89 \\
        & Color & 1.00 & 0.96 & 1.00 & 0.97 & 1.00 & 1.00 \\
        \bottomrule 
        
    \end{tabular}
    \label{tab:translation_results}
\end{table*}

\begin{table*}[!h]
\scriptsize
    \centering
    \caption{Example annotator comments suggesting corrections to translations.}
    \begin{tabular}{l}
    \toprule
        \textbf{Comment} \\
        \midrule
         $\hookrightarrow$ Should use ``above'' instead of ``on'' \\
         $\hookrightarrow$ More appropriate to use "memories'' instead of "impressions" \\
         $\hookrightarrow$ should be `small' balls (remove `round', add `small') \\
         $\hookrightarrow$ Particle suggests that numbers are written ``using" sheet of paper, not ``on" it. \\
         $\hookrightarrow$ ``On the side'' is translated as ``next to''. \\
         $\hookrightarrow$ ``toile d' araignée" can be directly translated to ``cobweb" \\
     \bottomrule
    \end{tabular}
    \label{tab:translation_correction_comments}
\end{table*}

\begin{table*}[!h]
\scriptsize
    \centering
    \caption{Prompt information for probing multilingual behavior in LLaVA.}
    
    \begin{tabular}{llp{0.6\textwidth}}
        \toprule
        \multicolumn{1}{c}{\textbf{Prompt Type}} & \multicolumn{1}{c}{\textbf{Language}} & \multicolumn{1}{c}{\textbf{Prompt}} \\
        \midrule

        \multirow{4}{*}{\textbf{Roles}} & English & (user, assistant) \\
        & French & (utilisateur, assistant) \\
        & German & (Benutzer, Assistent) \\
        & ... & ... \\
        \midrule
        
\multirow{4}{*}{\textbf{System}} & English & A conversation between a user and an LLM-based AI assistant. The assistant gives helpful and honest answers. \\
& French & Une conversation entre un utilisateur et un assistant IA basé sur LLM. L'assistant donne des réponses utiles et honnêtes. \\
& German & Ein Gespräch zwischen einem Benutzer und einem auf LLM basierenden KI-Assistenten. Der Assistent gibt hilfreiche und ehrliche Antworten. \\
& ... & ... \\
\midrule

\multirow{4}{*}{\textbf{User Prompt}} & English & What is in this image? Answer in English. \\
& French & Qu'est-ce qu'il y a dans cette image? Répondez en français. \\
& German & Was ist auf diesem Bild? Antwort auf Deutsch. \\
& ... & ... \\
\bottomrule

\end{tabular}
    \label{tab:llava-prompting}
\end{table*}

\clearpage

\begin{table*}[!h]
\scriptsize
    \centering
    
    \caption{A few examples of captions collected from the human study across English and Japanese speakers show differences in the observed content for each image. Japanese captions tend to include more context (e.g., background objects, added details). Samples are selected but representative of broader trends.}

    
    \begin{tabular}{cclp{8.25cm}}
        \toprule
        
        \raisebox{-1\height}[0pt][0pt]{\includegraphics[width=2cm]{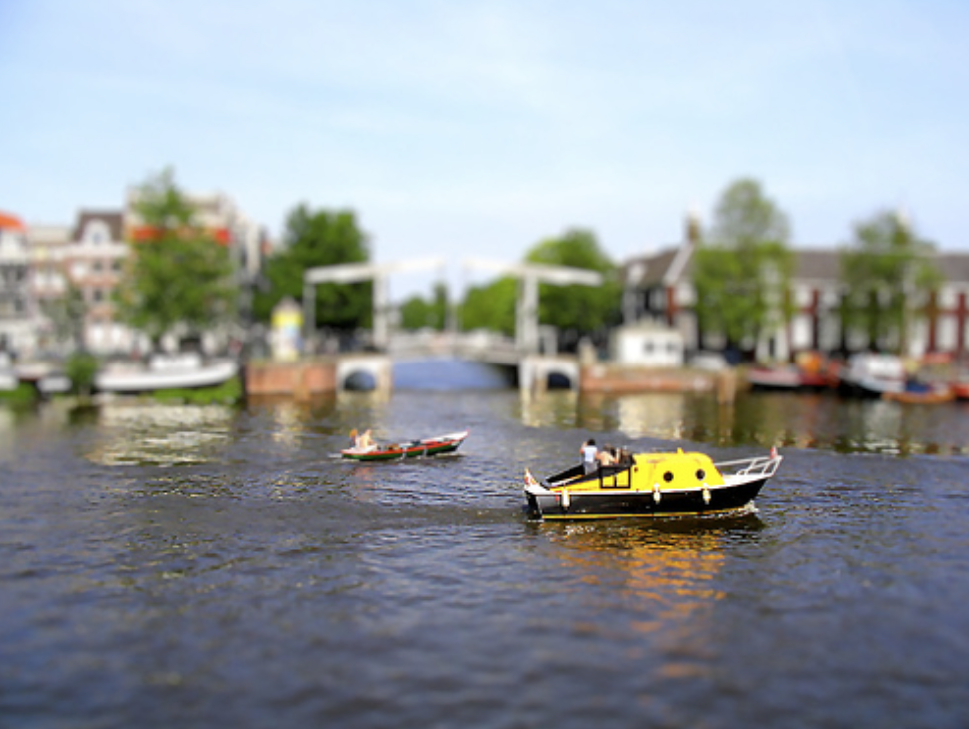}} & \multirow{3}{*}{English} 
         & {I.} & Two very small boats on a river  \\
         & & {II.} & Toy boats in the water \\
         & & {III.} & A yellow boat and a red boat that appear to be models. \\
         \cdashline{2-4}[1pt/2pt]\noalign{\vskip 1pt}
         & \multirow{3}{*}{Japanese} & 
         {I.} & Two boats on the water and a building in the back \\
         & & {II.} & close-up of a model of a boat and people on the waterfront \\
         & & {III.} & Two boats floating on the river and a model of the town in the distance \\
         
         \midrule

         \raisebox{-1\height}[0pt][0pt]{\includegraphics[width=2cm]{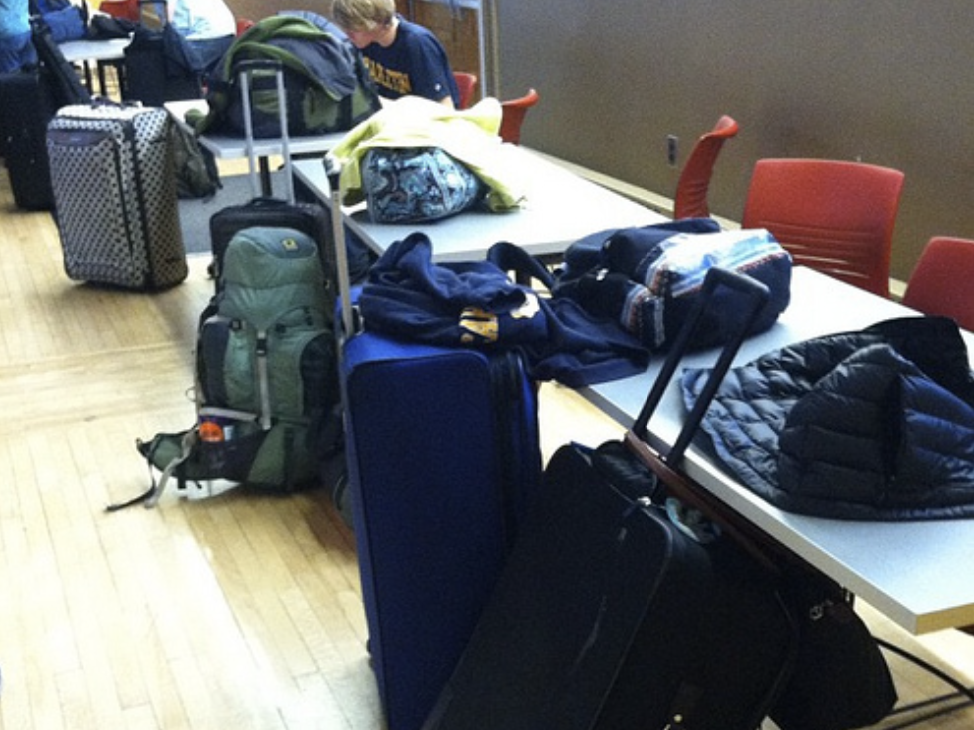}} & \multirow{3}{*}{English} & {I.} & Luggage left unattended at a table.  \\
         & & {II.} & Luggage lined up next to tables with jackets resting on the tables. \\
         & & {III.} & luggage sitting next to tables \\
         \cdashline{2-4}[1pt/2pt]\noalign{\vskip 1pt}
         & \multirow{3}{*}{Japanese} & {I.} & A man sitting in a lobby with lots of suitcases and bags \\
         & & {II.} & A man is sitting in a room, and there are several tables filled with luggage nearby. \\
         & & {III.} & Man waiting with a lot of luggage \\

         \midrule

         \raisebox{-1\height}[0pt][0pt]{\includegraphics[width=2cm]{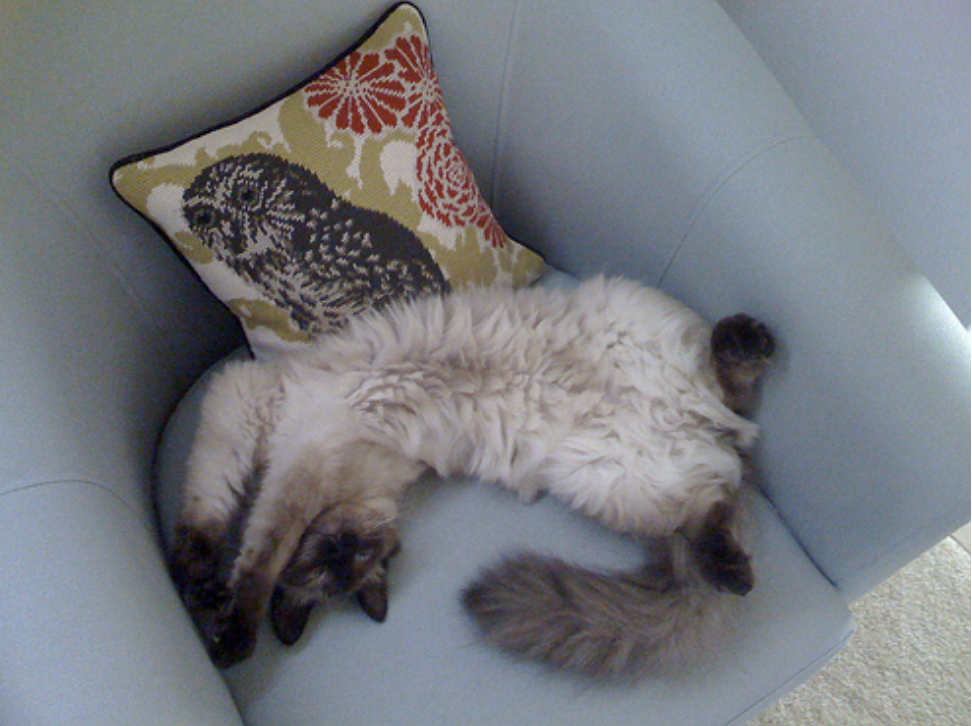}} & \multirow{3}{*}{English} 
         & {I.} & Cat laying down in an arm chair.  \\
         & & {II.} & A Siamese cat laying on its back on a couch next to a pillow. \\
         & & {III.} & A cat stretched out and upside down on a chair \\
         \cdashline{2-4}[1pt/2pt]\noalign{\vskip 1pt}
         & \multirow{3}{*}{Japanese} 
         & {I.} & A cat stretches out on a blue chair and a pillow with an embroidered owl next to it. \\
         & & {II.} & A cat is relaxing next to a cushion with a picture of an owl on it. \\
         & & {III.} & Cat sitting on his back in an armchair with an owl-patternedcushion \\
         
         \bottomrule
    \end{tabular}
    \label{fig:user-study-examples}
\end{table*}


\begin{sidewaystable*}
\scriptsize
\centering
\caption{Primary results across each of four evaluations. Shaded comparisons emphasize comparison between monolingual English caption sets and multilingual caption sets (including English and other languages), which have higher coverage/diversity across their respect evaluations.}
\label{primary-table}

\begin{tabular}{|c|c|c|c|r|r|r|r|r|r|r|r|r|r|r|}
\hline
\multirow{2}{*}{\textbf{Level}} & \multirow{2}{*}{\textbf{Evaluation}} & \multirow{2}{*}{\textbf{Data Source}} & \multirow{2}{*}{\textbf{Metric}} & \multicolumn{7}{c|}{\textbf{Monolingual}} & \multicolumn{4}{c|}{\textbf{Multilingual}} \\ 
\cline{5-15}
& & & & \textbf{en} & \textbf{fr} & \textbf{de} & \textbf{ru} & \textbf{zh} & \textbf{ja} & \textbf{ko} & \textbf{en,fr,zh} & \textbf{fr,zh,ru} & \textbf{de,fr,ru} & \textbf{avg}\footnote{The `avg' column is computed by randomly sampling across all languages.} \\ 
\hline
\multirow{21}{*}{Language} & \multirow{6}{*}{Logical meaning} & \multirow{3}{*}{Vertex} & Objects & 3.65 & 3.51 & 3.60 & 3.86 & 3.46 & 3.13 & 3.18 & 4.13 & 4.24 & 4.13 & 4.17 \\
\cline{4-15}
& & & Relations & 2.96 & 2.83 & 2.89 & 3.20 & 2.68 & 2.37 & 2.47 & 3.45 & 3.48 & 3.38 & 3.40 \\
\cline{4-15}
& & & Attributes & 1.67 & 1.67 & 1.79 & 1.86 & 1.66 & 1.59 & 1.62 & 2.29 & 2.40 & 2.33 & 2.33 \\
\cline{3-15}
& & \multirow{3}{*}{LLaVA} & Objects & 4.54 & 5.05 & 5.26 & 4.52 & 4.54 & - & \footnote{Japanese and Korean are excluded from LLaVA results due to extremely poor caption generation in the languages.} - & 6.14 & 6.15 & 6.25 & 5.93 \\
\cline{4-15}
& & & Relations & 3.79 & 4.21 & 4.42 & 3.67 & 3.66 & - & - & 4.85 & 4.76 & 4.97 & 4.54  \\
\cline{4-15}
& & & Attributes & 2.75 & 3.47 & 3.50 & 2.76 & 3.25 & - & - & 4.00 & 3.99 & 4.07 & 3.86 \\
\cline{2-15}
& & \multirow{3}{*}{XM} & Objects & 2.59 & 2.92 & 3.16 & 3.03 & 2.99 & 3.41 & 2.71 & 3.71 & 3.92 & 3.93 & 4.35 \\
\cline{4-15}
& & & Relations & 1.54 & 1.76 & 1.94 & 1.88 & 1.71 & 1.99 & 1.59 & 2.41 & 2.57 & 2.57 & 2.94 \\
\cline{4-15}
& & & Attributes & 1.27 & 1.66 & 1.97 & 1.74 & 2.01 & 2.47 & 1.46 & 2.36 & 2.59 & 2.55 & 2.97 \\
\hline
& \multirow{15}{*}{Expression} & \multirow{5}{*}{Vertex} & Concreteness & 1.64 & 1.64 & 1.67 & 1.66 & 1.51 & 1.50 & 1.56 & 1.75 & 1.74 & 1.73 & 1.81 \\
\cline{4-15}
& &  & Analytic & 0.85 & 0.43 & 0.62 & 1.08 & 2.3 & 2.6 & 2.17 & 2.02 & 2.05 & 1.0 & 2.3 \\
\cline{4-15}
& & & Clout & 4.94 & 5.05 & 5.40 & 7.14 & 6.92 & 6.49 & 5.56 & 10.59 & 11.29 & 9.8 & 11.01 \\
\cline{4-15}
& & & Authentic & 23.21 & 22.8 & 21.68 & 23.07 & 23.51 & 25.01 & 21.67 & 40.16 & 36.94 & 31.85 & 38.06 \\
\cline{4-15}
& & & Tone & 1.85 & 1.84 & 2.03 & 2.63 & 2.05 & 1.96 & 2.05 & 3.98 & 4.10 & 4.19 & 3.92 \\
\cline{3-15}
& & \multirow{5}{*}{LLaVA} & Concreteness & 2.17 & 2.44 & 2.53 & 2.27 & 2.33 & - & - & 2.54 & 2.54 & 2.57 & 2.56 \\
\cline{4-15}
& &  & Analytic & 2.85 & 4.64 & 14.72 & 7.53 & 23.05 & - & - & 22.03 & 22.29 & 23.76 & 19.61 \\
\cline{4-15}
& & & Clout & 14.19 & 19.96 & 15.81 & 23.15 & 21.3 & - & - & 23.06 & 26.78 & 26.84 & 28.72 \\
\cline{4-15}
& & & Authentic & 35.97 & 38.01 & 34.33 & 37.20 & 44.63 & - & - & 54.94 & 54.64 & 54.82 & 53.15 \\
\cline{4-15}
& & & Tone & 6.79 & 9.53 & 16.64 & 16.48 & 11.56 & - & - & 16.74 & 20.79 & 19.80 & 16.56 \\
\cline{3-15}
& & \multirow{5}{*}{XM} & Concreteness & 1.80 & 2.24 & 2.11 & 2.03 & 2.26 & 2.25 & 2.14 & 2.08 & 2.13 & 2.10 & 2.17\\
\cline{4-15}
& &  & Analytic & 1.88 & 2.31 & 1.62 & 0.76 & 5.02 & 5.28 & 5.13 & 4.12 & 4.37 & 4.47 & 4.49 \\
\cline{4-15}
& & & Clout & 9.66 & 14.77 & 15.00 & 14.14 & 13.32 & 13.10 & 12.24 & 18.8 & 20.51 & 19.54 & 19.39 \\
\cline{4-15}
& & & Authentic & 33.21 & 40.09 & 42.22 & 32.35 & 34.02 & 33.83 & 35.28 & 53.12 & 54.41 & 53.35 & 53.21 \\
\cline{4-15}
& & & Tone & 8.62 & 12.16 & 9.74 & 10.90 & 10.59 & 9.18 & 9.07 & 13.78 & 14.69 & 15.42 & 15.40 \\
\hline

\multirow{6}{*}{Model} & \multirow{3}{*}{Embeddings} & Vertex & \multirow{3}{*}{Coverage} & 0.19 & 0.18 & 0.19 & 0.22 & 0.20 & 0.19 & 0.37 & 0.37 & 0.33 & 0.38 & 0.49 \\
\cline{3-3}
\cline{5-15}
& & LLaVA & & 0.22 & 0.29 & 0.28 & 0.29 & 0.36 & - & - & 0.45 & 0.47 & 0.43 & 0.47 \\
\cline{3-3}
\cline{5-15}
& & XM & & 0.38 & 0.0.42 & 0.43 & 0.40 & 0.46 & 0.42 & 0.43 & 0.54 & 0.54 & 0.49 & 0.52  \\
\cline{2-15}

& \multirow{3}{*}{Fine-tuning} & Vertex & \multirow{3}{*}{Best Model}\footnote{Each cell value is the original language of the caption set used to train the model which performed best as evaluated by the caption set originally from the language specified in the column.} & en & de & fr & ru & zh & ja & ko & - & - & - & multi \\
\cline{3-3}
\cline{5-15}
& & LLaVA & & en & de & fr & ru & zh & - &  - & - & - & - & multi \\
\cline{3-3}
\cline{5-15}
& & XM & & en & de & fr & ru & zh & ja & ko & - &-  &- & multi \\
\hline
\end{tabular}
\end{sidewaystable*}

\clearpage

\begin{table*}[p]
\scriptsize
    \centering
    \caption{Sizes of intersections between monolingual unioned scene graphs, listed in the form ``number of objects / number of relations''. Sizes listed along the diagonal arc of monolingual graphs and can be used for reference.}
    \begin{tabular}{c|ccccccc}
    \toprule
        & en & de & fr & ru & zh & ja & ko \\
    \midrule
        en & 3.65 / 2.96 & 2.34 / 1.20 & 2.41 / 1.26 & 2.39 / 1.24 & 2.15 / 0.97 & 2.04 / 0.91 & 2.02 / 0.89 \\
        de & & 3.51 / 2.83 & 2.37 / 1.24 & 2.47 / 1.32 & 2.14 / 0.98 & 2.04 / 0.91 & 2.06 / 0.94 \\
        fr & & & 3.60 / 2.89 & 2.44 / 1.27 & 2.16 / 0.97 & 2.07 / 0.91 & 2.09 / 0.95 \\
        ru & & & & 3.86 / 3.2 & 2.25 / 1.06 & 2.13 / 0.98 & 2.12 / 0.96 \\
        zh & & & & & 3.46 / 2.68  & 2.08 / 0.95 & 2.04 / 0.92 \\
        ja & & & & & & 3.13 / 2.37 & 2.10 / 1.02 \\
        ko & & & & & & & 3.18 / 2.47 \\
    \bottomrule
    \end{tabular}
    \label{tab:crossling_metrics}
\end{table*}

\begin{table*}[p]
    \scriptsize
    \centering
    \caption{Scene graph metrics across Vertex and LLaVA captions in different languages show that multilingual scene graph unions are richer than monolingual ones. Increases are relative to the English average.}
    \begin{tabular}{llccc}
    \toprule
     & & \textbf{en,fr,zh} & \textbf{fr,de,ru} & \textbf{multi-model} \\
    \midrule
    \multirow{3}{*}{\textbf{Vertex}} & \textbf{Objects} & 4.31 & 4.25 & 4.63 \\
                                    & \textbf{Relations} & 3.60 & 3.56 & 3.64 \\
                                    & \textbf{Attributes} & 2.13 & 2.15 & 2.19 \\
    \midrule
    \multirow{3}{*}{\textbf{LLaVA}} & \textbf{Objects} & 5.87 & 6.02 & 6.65 \\
                                    & \textbf{Relations} & 4.84 & 4.97 & 5.42 \\
                                    & \textbf{Attributes} & 4.10 & 4.07 & 2.88 \\
    \bottomrule
    \end{tabular}
    \label{tab:multimodel-comparison}
\end{table*}

\begin{table*}[p]
\scriptsize
\def\arraystretch{1}
    \centering
    \caption{Intersection sizes between 3 unioned monolingual Vertex captions and an English multimodel baseline (a unioned BLIP2 $\cup$ GIT scene graph, held constant across all languages) are both relatively \textbf{small} and \textbf{smaller for Asian than European languages}. All relationships between \one{European languages} and \two{Asian languages} are statistically significant with Bonferroni correction. The `mm' column includes the size of the unioned GIT and BLIP model scene graph for reference.}
    \begin{tabular}{cc ccccccc c}
        \toprule
        & \multicolumn{1}{c}{} & \multicolumn{7}{c}{\textbf{Language}} \\
        \cmidrule{3-9}
        & & \textbf{en} & \textbf{de} & \textbf{fr} & \textbf{ru} & \textbf{zh} & \textbf{ja} & \textbf{ko} & \textbf{mm}  \\
        \cmidrule{2-10}
        & \textbf{Objects} & \one{1.96} & \one{1.92} & \one{1.93} & \one{1.97} & \two{1.85} & \two{1.73} & \two{1.76} & 3.59 \\
        & \textbf{Relations} & \one{0.79} & \one{0.76} & \one{0.74} & \one{0.78} & \two{0.70} & \two{0.62} & \two{0.64} & 2.51 \\
        \rot{\rlap{\textbf{Metric}}}
        & \textbf{Attributes} & 0.44 & 0.36 & 0.37 & 0.42 & 0.37 & 0.37 & 0.33 & 1.45 \\
        
        \bottomrule
    \end{tabular}
    \label{tab:multimodel-intersections}
\end{table*}

\clearpage

\begin{table*}[p]
\footnotesize
    \centering
    \caption{Evaluations for models fine-tuned on LLaVA captions. Generally speaking, a model fine-tuning on a particular language performs best on that language.}
    \begin{tabular}{cr|cccccc}
        \toprule
        & \multicolumn{1}{c}{} & \multicolumn{6}{c}{\textbf{Evaluated on}}\\
         & & \textbf{en} & \textbf{de} & \textbf{fr} & \textbf{ru} & \textbf{zh} & \textbf{multi} \\
        \cmidrule{2-8}
        & \textbf{en} & \first{0.271} & 0.225 & 0.229 & 0.219 & 0.218 & 0.230 \\
        & \textbf{de} & 0.213 & \first{0.245} & 0.219 & 0.217 & 0.215 & 0.219 \\
        & \textbf{fr} & 0.248 & 0.240 & \first{0.259} & 0.234 & 0.236 & 0.246 \\
        & \textbf{ru} & 0.226 & 0.234 & 0.228 & \first{0.254} & 0.231 & 0.239 \\
        \rot{\rlap{\textbf{Fine-tuned on}}}
        & \textbf{zh} & 0.199 & 0.202 & 0.199 & 0.207 & \first{0.247} & 0.216 \\
        & \textbf{multi} & \second{0.239} & \second{0.233} & \second{0.234} & \second{0.233} & \second{0.235} & \second{0.244} \\
        \bottomrule
    \end{tabular}
    \label{tab:finetuning_llava}
\end{table*}

\begin{table*}[p]
\footnotesize
    \centering
    \caption{Evaluations for models fine-tuned on XM captions. Generally speaking, a model fine-tuning on a particular language performs best on that language.}
    \def\arraystretch{1.15}
    \begin{tabular}{cr|cccccccc}
        \toprule
     & \multicolumn{1}{c}{} & \multicolumn{8}{c}{\textbf{Evaluated on}}
        \\
         & & \textbf{en} & \textbf{de} & \textbf{fr} & \textbf{ru} & \textbf{zh} & \textbf{ja} & \textbf{ko} & \textbf{multi} \\
        \cmidrule{2-10}
        & \textbf{en} & \first{0.254} & 0.124 & 0.1421 & 0.120 & 0.114 & 0.129 & 0.130 & 0.148 \\
        & \textbf{de} & 0.158 & \first{0.153} & 0.152 & 0.143 & 0.124 & 0.140 & 0.146 & 0.149 \\
        & \textbf{fr} & 0.182 & 0.142 & \first{0.181} & 0.143 & 0.130 & 0.146 & 0.150 & 0.154 \\
        & \textbf{ru} & 0.172 & 0.136 & 0.152 & \first{0.159} & 0.125 & 0.137 & 0.142 & 0.148 \\
        & \textbf{zh} & 0.144 & 0.116 & 0.129 & 0.120 & \first{0.124} & 0.130 & 0.142 & 0.130 \\
        \rot{\rlap{\textbf{~Fine-tuned on}}}
        & \textbf{ja} & 0.144 & 0.128 & 0.137 & 0.125 & 0.124 & \first{0.154} & 0.144 & 0.135 \\
        
        & \textbf{ko} & 0.151 & 0.116 & 0.131 & 0.116 & 0.115 & 0.134 & \first{0.159} & 0.134 \\
        & \textbf{multi} & \second{0.179} & \second{0.140} & \second{0.153} & \second{0.145} & \second{0.131} & \second{0.149} & \second{0.151} & \second{0.151} \\
        \bottomrule
    \end{tabular}
    \label{tab:finetuning-xm}
\end{table*}


\newpage

\begin{table*}[p]
\scriptsize
    \centering
    \caption{Examples in which multilingual distributions identify visual features which are not documented in the Visual Genome dataset. Rightmost column indicates objects mentioned in multilingual scene graphs but which are not covered in the Visual Genome object list, shown in the left column.}
    \begin{tabular}{lp{5cm}p{4cm}}
         \toprule
         \textbf{Image} & \textbf{VG Objects} & \textbf{Scene Graph Objects} \\
         \midrule
         \raisebox{-1\height}[0pt][0pt]{\includegraphics[width=3cm]{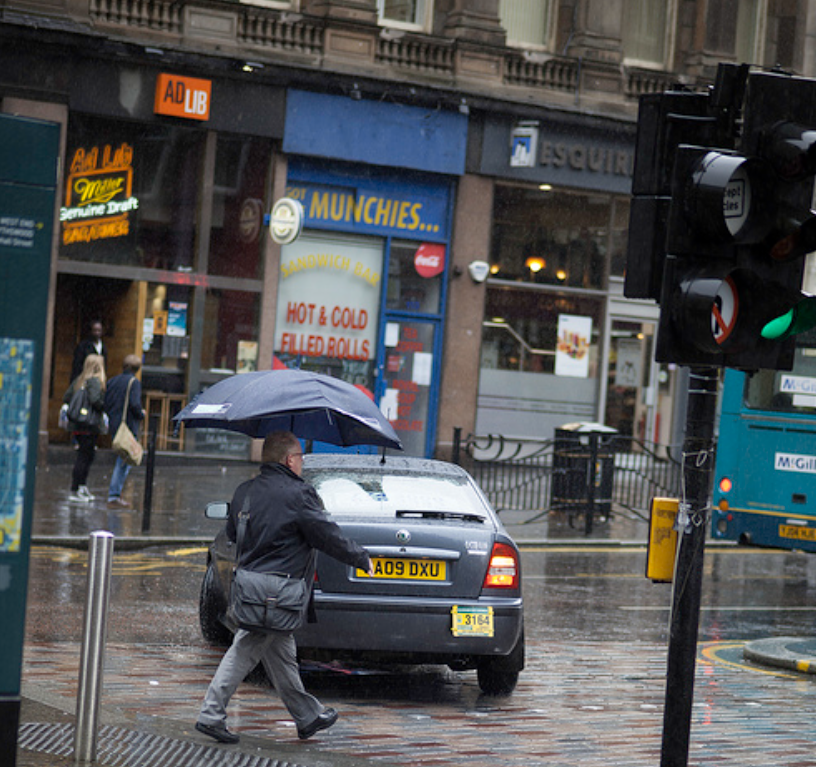}} & \texttt{woman, sign, man, bag, license plate, car, person, leg, satchel} & \texttt{umbrella, sandwich restaurant, street, rain} \vspace{25mm} \\
         \midrule
         \raisebox{-1\height}[0pt][0pt]{\includegraphics[width=3cm]{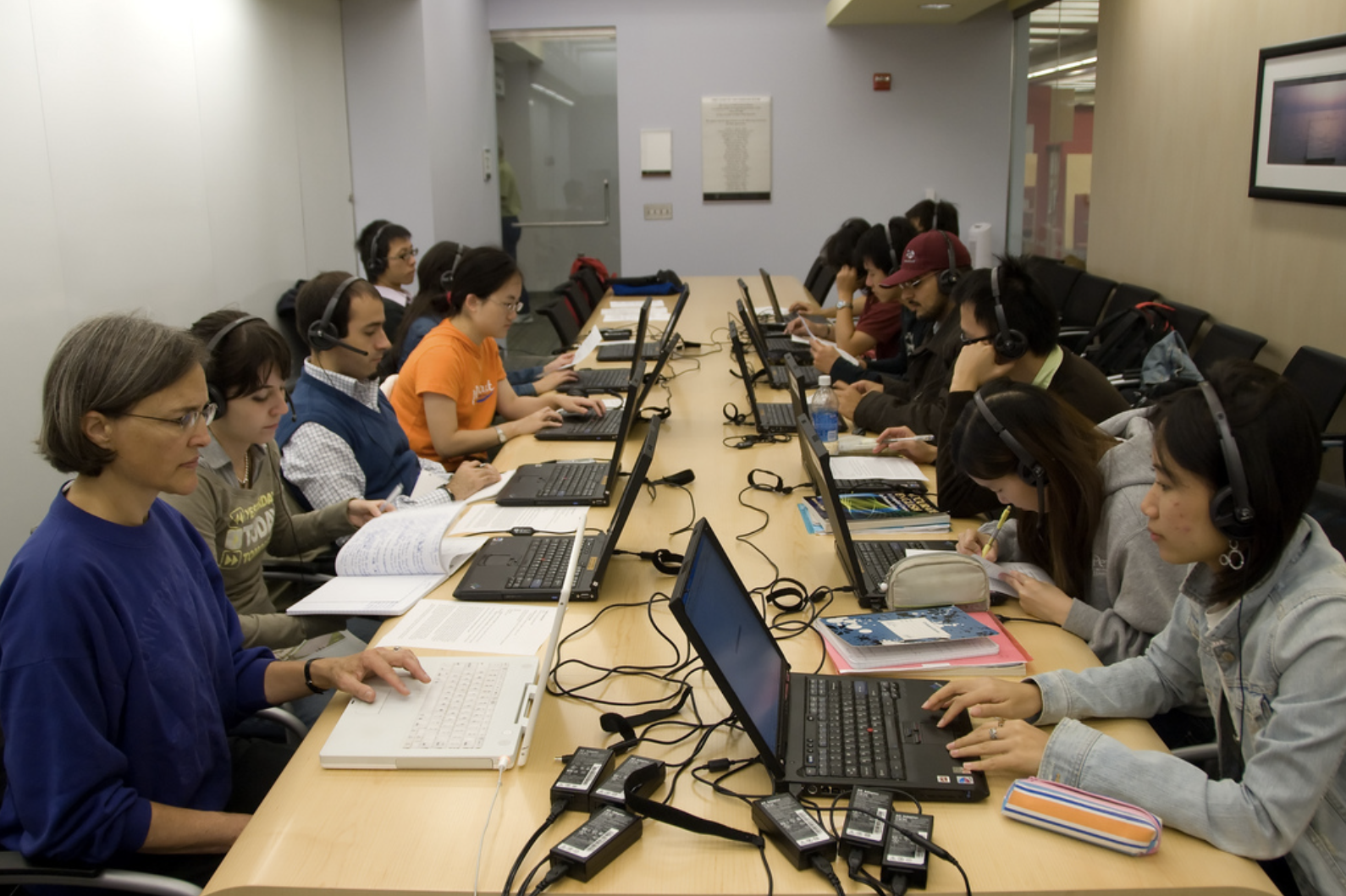}} & \texttt{woman, key, notes, page, keyboard, pencil case, laptop, student} & \texttt{table} \vspace{20mm} \\
         \midrule
         \raisebox{-1\height}[0pt][0pt]{\includegraphics[width=3cm]{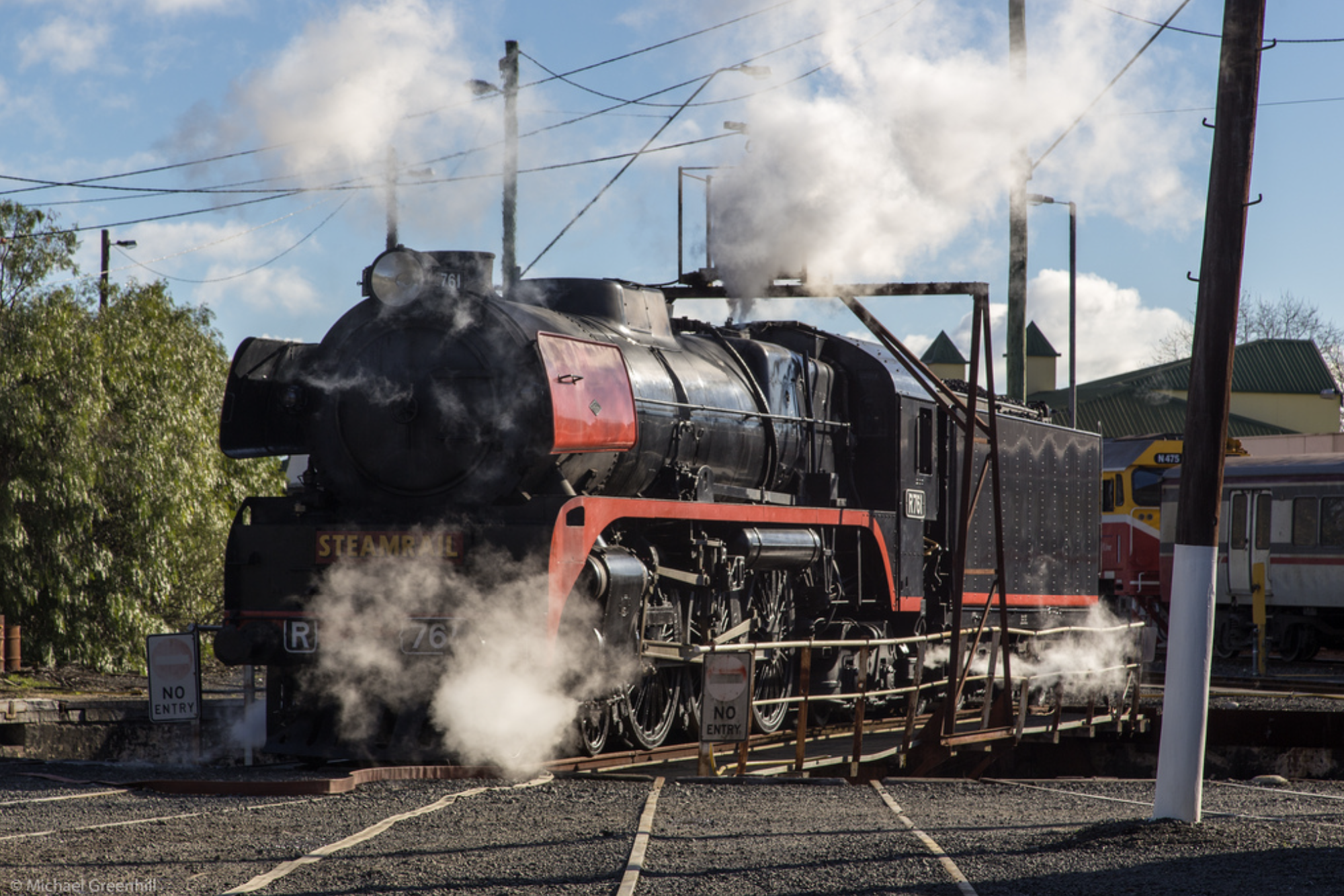}} & \texttt{leaves, sign, sky, cloud, trees, roof, train, steam cloud, ground, lamp, green leaves, cables, pole, tracks, locomotive, train car, tree, steeples, gravel, steam, bush, door, wheel} & \texttt{number, logo, inscription} \vspace{20mm} \\
         \midrule
         \raisebox{-1\height}[0pt][0pt]{\includegraphics[width=3cm]{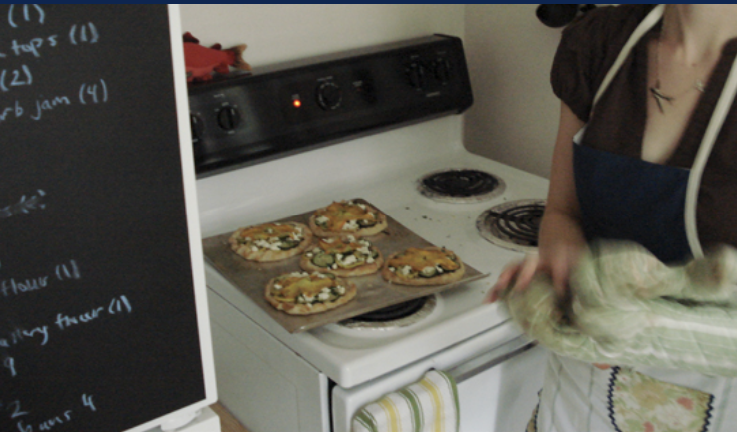}} & \texttt{tray, writing, cloth, stove door, light, oven back, bird necklace, mitt, shirt, apron, stove, burner, strings, aprontop, towel, board, pizza, shortsleeveshirt, menu, woman, necklace, pizzas, pan, oven, sheet} & \texttt{chalkboard} \vspace{20mm} \\
         \midrule
         \raisebox{-1\height}[0pt][0pt]{\includegraphics[width=3cm]{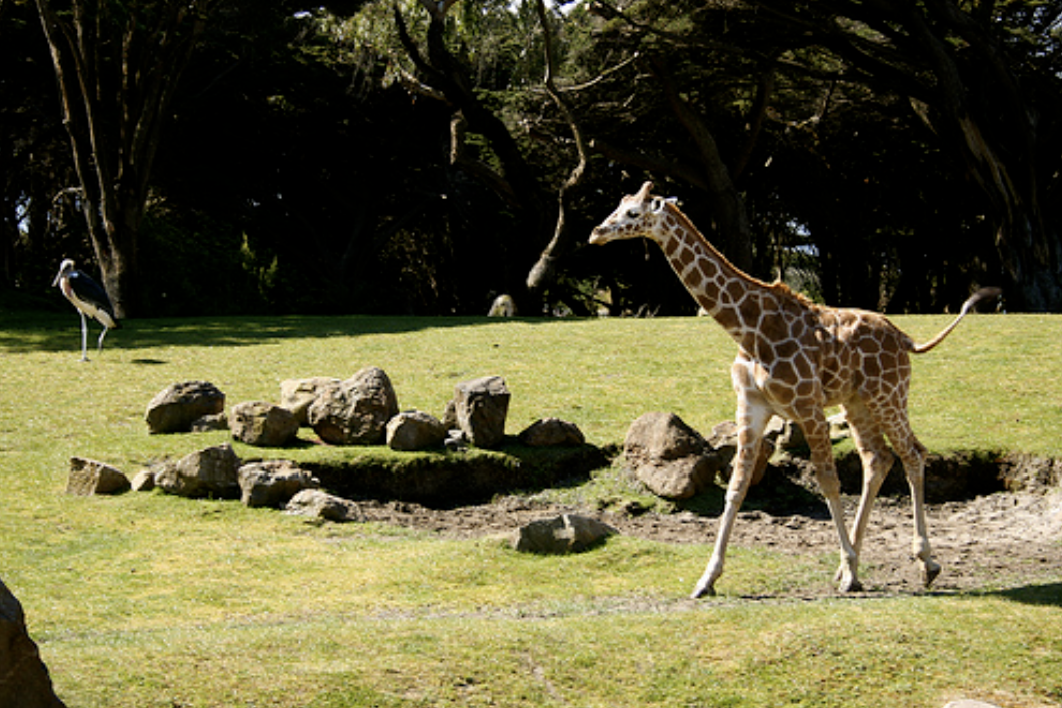}} & \texttt{giraffe tail, spot, rock, giraffe, rocks, grass} & \texttt{bird} (left of image) \vspace{20mm} \\
         \bottomrule
    \end{tabular}
    \label{tab:vg-exceed-examples}
\end{table*}

\begin{figure*}[p]
    \centering
    \begin{subfigure}[b]{\textwidth}
        \includegraphics[width=\textwidth]{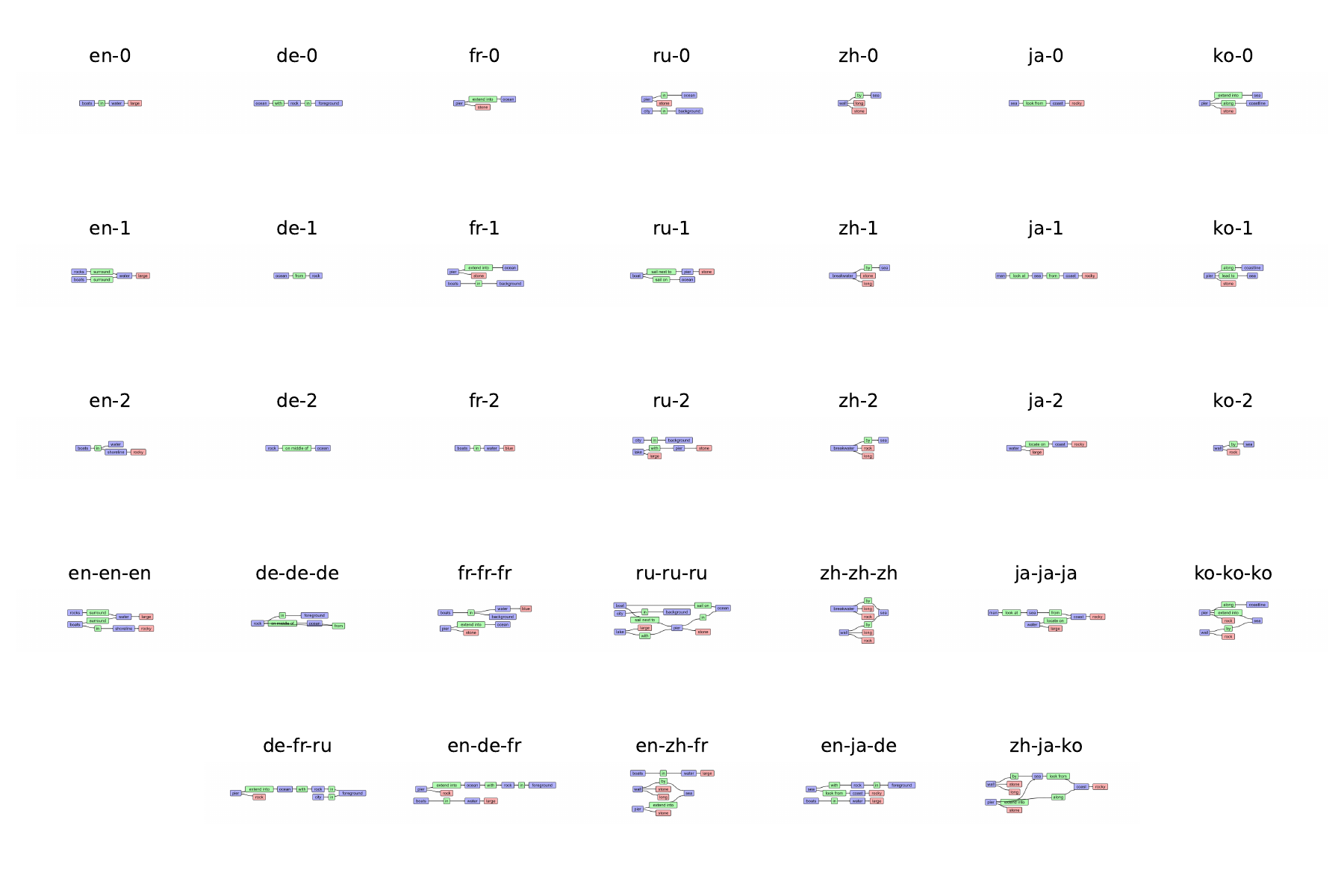}
    \end{subfigure}
    \begin{subfigure}[b]{\textwidth}
        \includegraphics[width=\textwidth]{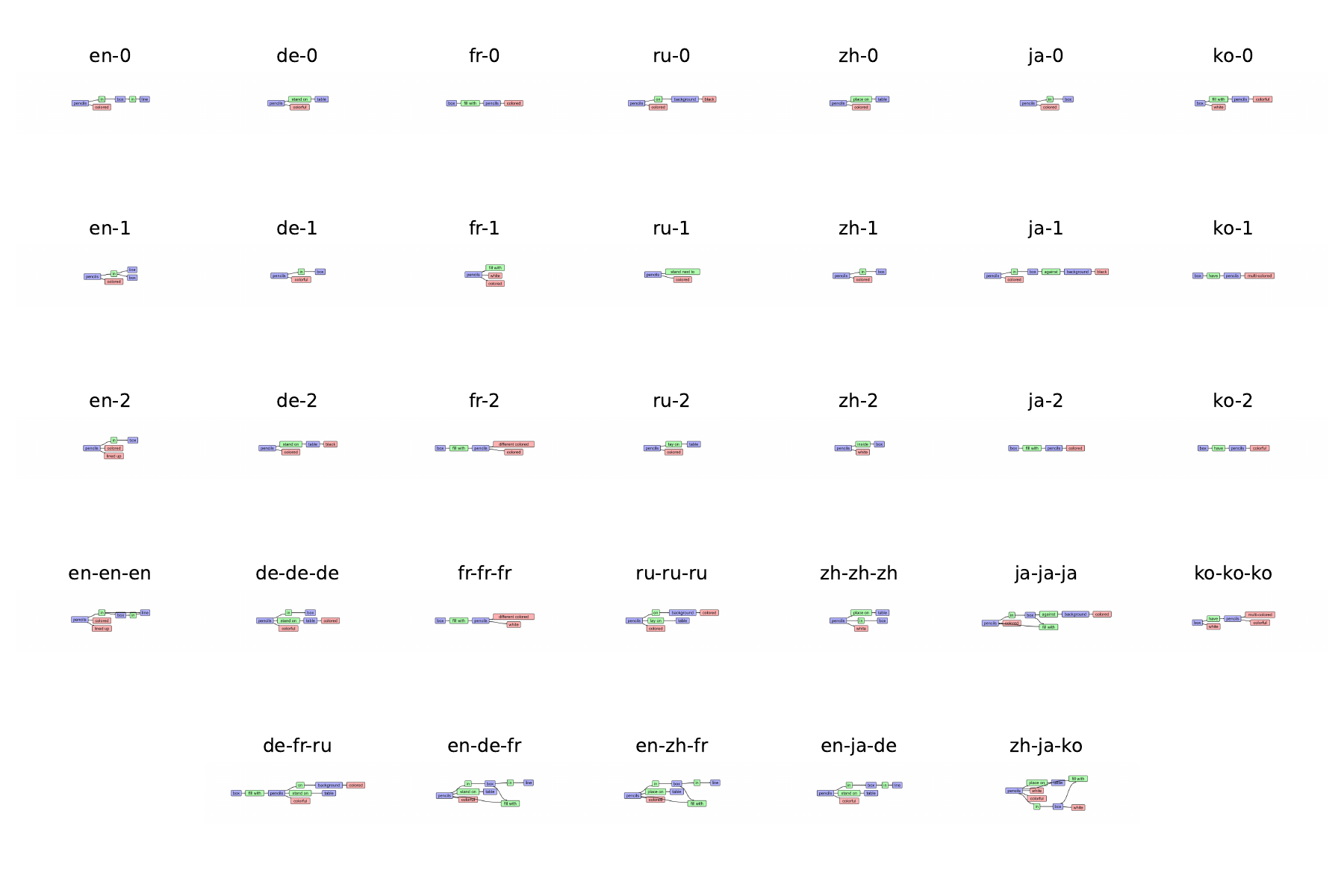}
    \end{subfigure}
    \caption{Sample scene graphs across six images. ``lang-$n$'' indicates the scene graph generated for the $n$th caption in lang. ``lang1-lang2-lang3'' indicates the scene graph unioned from three scene graphs originally from each of the three languages.}
    \label{fig:big-scene-graph-spread}
\end{figure*}
\begin{figure*}[p]
    \ContinuedFloat
    \begin{subfigure}[b]{\textwidth}
        \includegraphics[width=\textwidth]{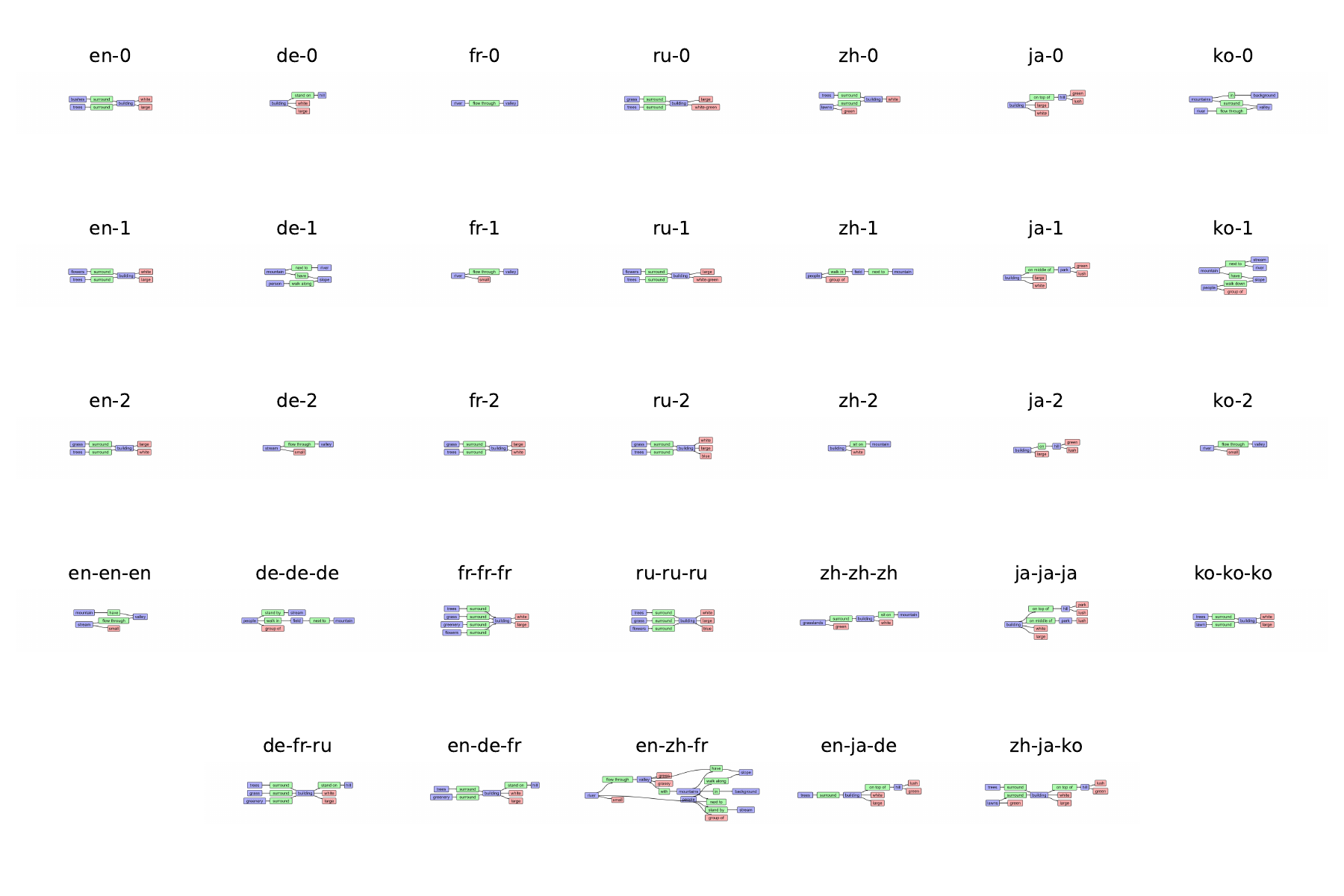}
    \end{subfigure}
    \begin{subfigure}[b]{\textwidth}
        \includegraphics[width=\textwidth]{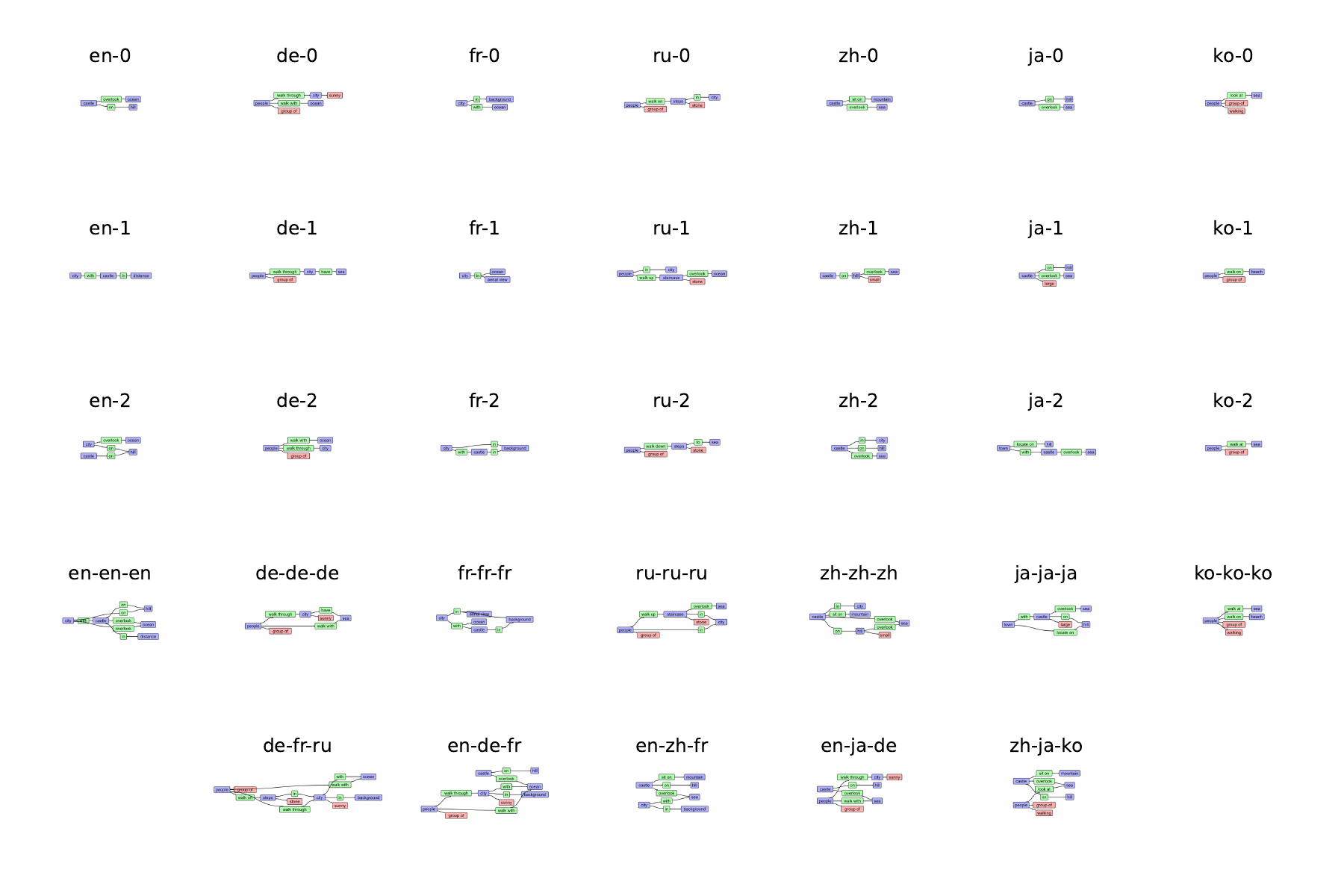}
    \end{subfigure}
\end{figure*}
\begin{figure*}[p]
    \ContinuedFloat
    \begin{subfigure}[b]{\textwidth}
        \includegraphics[width=\textwidth]{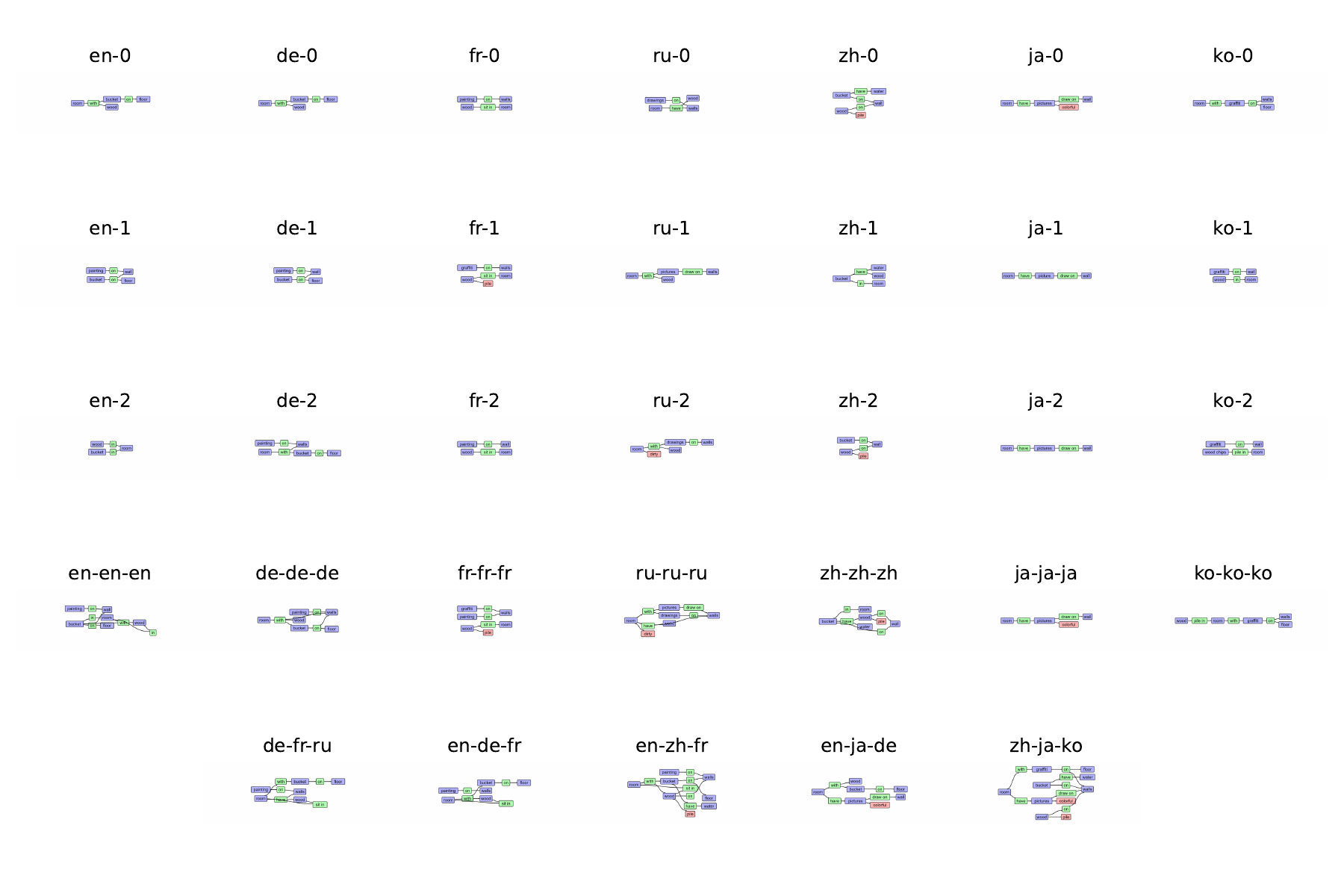}
    \end{subfigure}
    \begin{subfigure}[b]{\textwidth}
        \includegraphics[width=\textwidth]{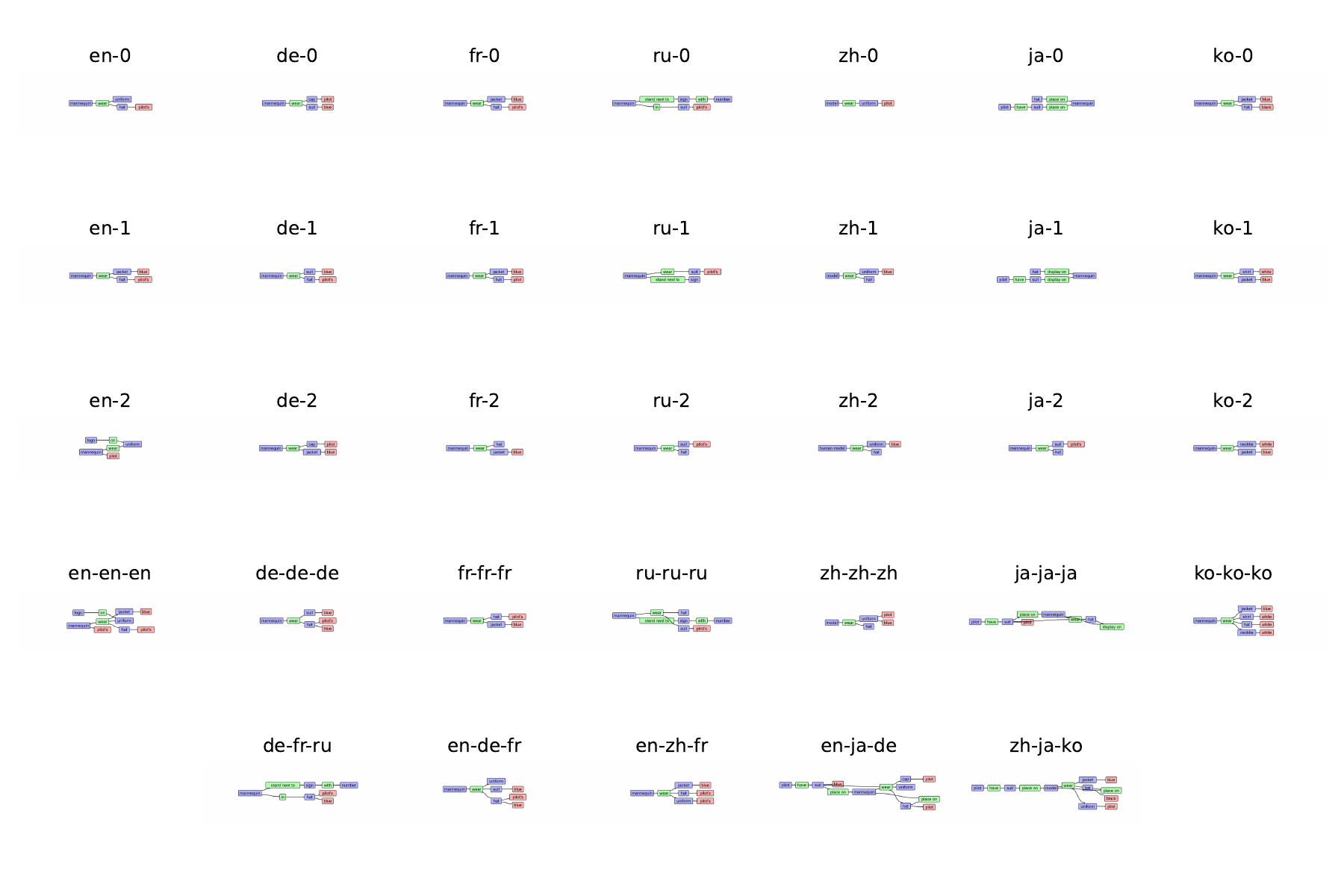}
    \end{subfigure}
\end{figure*}

\end{document}